\documentclass[review]{elsarticle}

\usepackage{lineno,hyperref}
\modulolinenumbers[5]

\usepackage{amsmath,amssymb,amsfonts}
\usepackage{algorithmic}
\usepackage{graphicx}
\usepackage{textcomp}
\usepackage{xcolor}

\usepackage{fancyhdr}

\usepackage[utf8]{inputenc}		
\usepackage{color}				
\usepackage{microtype} 			
\usepackage{array}
\newcolumntype{C}[1]{>{\centering\arraybackslash}p{#1}}
\newcolumntype{L}[1]{>{\raggedright\arraybackslash}p{#1}}
\newcolumntype{R}[1]{>{\raggedleft\arraybackslash}p{#1}}
\usepackage{arydshln}

\journal{Expert Systems With Applications}

\biboptions{authoryear}

\begin{document}

\begin{frontmatter}

\title{A white-box analysis on the writer-independent dichotomy transformation applied to offline handwritten signature verification}

\author[ufpe]{Victor L. F. Souza \corref{cor1}} 
\ead{vlfs@cin.ufpe.br}
\cortext[cor1]{Corresponding author}

\author[ufpe]{Adriano L. I. Oliveira}
\ead{alio@cin.ufpe.br}

\author[ets]{Rafael M. O. Cruz}
\ead{rafaelmenelau@gmail.com}

\author[ets]{Robert Sabourin}
\ead{robert.sabourin@etsmtl.ca}

\address[ufpe]{
    Centro de Inform\'atica, Universidade Federal de Pernambuco, Recife (PE), Brazil
}

\address[ets]{
    \'Ecole de Technologie Sup\'erieure - Universit\'e du Qu\'ebec, Montreal, Qu\'ebec, Canada
}

\begin{abstract}

High number of writers, small number of training samples per writer with high intra-class variability and heavily imbalanced class distributions are among the challenges and difficulties of the offline Handwritten Signature Verification (HSV) problem. A good alternative to tackle these issues is to use a writer-independent (WI) framework. In WI systems, a single model is trained to perform signature verification for all writers from a dissimilarity space generated by the dichotomy transformation. Among the advantages of this framework is its scalability to deal with some of these challenges and its ease in managing new writers, and hence of being used in a transfer learning context. In this work, we present a white-box analysis of this approach highlighting how it handles the challenges, the dynamic selection of references through fusion function, and its application for transfer learning. All the analyses are carried out at the instance level using the instance hardness (IH) measure. The experimental results show that, using the IH analysis, we were able to characterize ``good'' and ``bad'' quality skilled forgeries as well as the frontier region between positive and negative samples. This enables futures investigations on methods for improving discrimination between genuine signatures and skilled forgeries by considering these characterizations.

\end{abstract}


\end{frontmatter}


\section{Introduction}

Handwritten Signature Verification (HSV) systems are used to automatically recognize whether the signature provided by a writer belongs to the claimed person \citep{guru:17}. 
In offline HSV, the signature is acquired after the writing process is completed, and the system deals with the signature as an image.
For instance, credit card transactions or document authentication are among real-world applications using HSV systems \citep{hafemann_review:17, zois:19asymmetric}.

In the HSV problem, genuine signatures are the ones produced by the claimed person (original writer) and forgeries are those created by an impostor (forger). In general, forgeries can be categorized, based on the knowledge of the forger, into the following types \citep{masoudnia:19}:
\begin{itemize}
\item Random forgeries: the forger has no information about the original writer. 
\item Simple forgeries: the forger knows the name of the original writer, but does not have access to the signature pattern.
\item Skilled forgeries: the forger has information about both the name and the genuine signature pattern of the original writer, resulting in forgeries that are more similar to genuine signatures.
\end{itemize}

\begin{figure}[!htb]
\centering
  \includegraphics[width=\columnwidth]{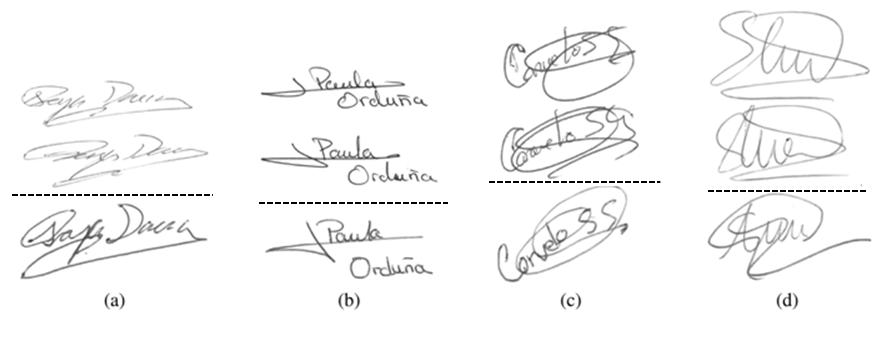}
  \caption{Signature examples from the GPDS dataset. Each column shows two genuine signatures from the same writer (above the line) and a skilled forgery (under the line).}
  \label{fig:signature_examples}
\end{figure}

Figure \ref{fig:signature_examples} depicts examples of genuine signatures and skilled forgeries, obtained from \citep{hafemann:17}. Each column shows two genuine signatures from the same writer and a skilled forgery, from the GPDS dataset. 

A first aspect that should be considered when working with HSV is the decision among which classification strategy to use, that is, WD vs. WI \citep{bouamra:18}.
If a verification model is trained for each writer, the system is called writer-dependent (WD). This approach is the most common and in general, achieves better classification accuracies. However, requiring a classifier for each writer increases the complexity, and the computational cost of the system operations as more writers are added \citep{eskander:13}.
On the other hand, in writer-independent (WI) systems, a single model is trained for all writers. In this scenario, the systems usually operate on the dissimilarity space generated by the dichotomy transformation \citep{rivard:13}. In this approach, a dissimilarity (distance) measure is used to compare samples (query and reference samples) as belonging to the same or another writer. \citep{eskander:13}.
When compared to the WD approach, WI systems are less complex, but in general obtain worse accuracy \citep{hafemann_review:17}.

\begin{figure}[!htb]
\centering
  \includegraphics[width=\columnwidth]{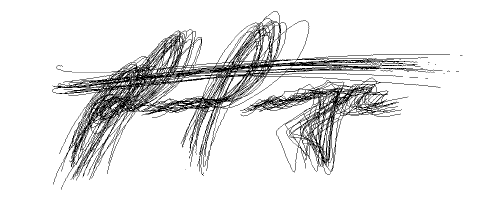}
  \caption{Overlaid genuine signature images of a single writer, presenting the intra-class variability of the data.}
  \label{fig:signature_skeleton}
\end{figure}

Some of the challenges related to the offline HSV are: ($C_1$) the high number of writers (classes), ($C_2$) the high-dimensional feature space, ($C_3$) small number of training samples per writer with high intra-class variability (Figure \ref{fig:signature_skeleton} shows an example of this problem in the genuine signatures) and ($C_4$) the heavily imbalanced class distributions \citep{hafemann_review:17, masoudnia:19}. 

Still, the main challenge is faced when dealing with skilled forgeries ($C_5$). Even though they are the most similar to genuine signatures, in general, they are not available for training purposes in real HSV applications. Thus, the systems are trained with partial knowledge as the classifiers are trained without sufficient information to distinguish between genuine signatures and skilled forgeries \citep{hafemann_review:17}.

The dichotomy transformation (DT) can be applied to deal with some of these challenges and therefore facilitate the signature verification task. The samples in the dissimilarity space (DS) generated by the dichotomy transformation are formed through the pairwise comparisons of signatures (a questioned and a reference signature) in the feature space \citep{souza:18}. Thus, having a good feature representation of the signatures is very important for DT to work \citep{bertolini:10}.

Also, as the classification only depends on the input reference signature, by using the DT in a writer-independent approach, the model can verify signatures of writers for whom the classifier was not trained. So, it can easily manage new incoming writers ($C_6$). 
In this way, WI systems have the advantages of being scalable and adaptable, and may even be used for transfer learning, which is a methodology that tries to use the knowledge acquired for one task to solve related ones \citep{shao:15}. 

The dichotomy transformation is a very important technique to solve some of the problems related to HSV problem, as demonstrated in two preliminary studies \citep{souza_dissimilarity:19} and \citep{souza_characterization:19}. In \citep{souza_dissimilarity:19}, we have shown that 
even DT increasing the number of samples in the offline WI-HSV context, as a consequence, redundant information is generated. Thus, prototype selection can be used to discard redundant samples without degrading the verification performance of the classifier when compared to the model trained with all available samples.
Furthermore, we have discussed how a WI-SVM trained in the GPDS can be used to verify signatures in other datasets without any further transfer adaptation in the WI-HSV context and still obtain similar results when compared to both WD and WI classifiers trained and tested in the own datasets. In \citep{souza_characterization:19} an instance hardness (IH) analysis showed the following behavior of samples in DS: while positive samples form a compact cluster located close to the origin, negative samples have a sparse distribution in the space.

However, in our preliminary studies \citep{souza_dissimilarity:19, souza_characterization:19}, some important aspects were not investigated, such as: 
\begin{enumerate}
    \item The use of a methodology for analysing the instance hardness (IH), considering the neighborhood of the training set of the GPDS dataset (instead of considering the test sets, as in \cite{souza_dissimilarity:19} and \cite{souza_characterization:19});
    \item The characterization of the dynamic reference selection using the MAX as fusion function;
    \item Analysis of the accuracy as a function of the IH in the GPDS and in transfer learning;
    \item A characterization of ``good'' and ``bad'' quality skilled forgeries;
    \item The overlap of positive and negative classes and the need for a complex decision function. 
\end{enumerate}

To fill this gap, this paper addresses all these issues and provides a white-box understanding of DT applied in a WI framework for handwritten signature verification. 
All the analyses are carried out based on the instance hardness (IH) measure \citep{smith:14} to maintain the findings at the instance level \citep{lorena:18}.
The performed experiments aim to allow future works on better methods for classification of good quality skilled forgeries.

This paper is organized as follows: Section \ref{sec:basic_concepts} contains the basic concepts related to this work, which are the dichotomy transformation applied to the HSV context, the used feature representation, the transfer learning and the instance hardness. Section \ref{sec:overview} the overview of the proposed approach. Section \ref{sec:experiments} contains the experiments and the discussion about the obtained results. Section \ref{sec:lessons_learned} presents the lessons learned in the study. Conclusions and future works are presented in the last section.

\section{Basic concepts}
\label{sec:basic_concepts}

\subsection{WI dichotomy transformation for handling HSV data difficulties}
\label{sec:dichotomy_transformation}

The Dichotomy Transformation (DT) approach \cite{cha:00}, allows to transform a multi-class pattern recognition problem into a $2$-class problem. 
In this approach, a dissimilarity (distance) measure is used to determine whether a given reference signature and a questioned signature belonging to the same writer \citep{eskander:13}.

Formally, let $\textbf{x}_q$ and $\textbf{x}_r$ be two feature vectors in the feature space, the dissimilarity vector resulting from the Dichotomy Transformation, $\textbf{u}$, is computed by equation \ref{eq:DT_distance}:

\begin{equation}
\label{eq:DT_distance}
  \textbf{u}(\textbf{x}_{q},\textbf{x}_{r}) =
  \begin{bmatrix}
    |x_{q1} - x_{r1}| \\
    |x_{q2} - x_{r2}| \\
    \vdots \\
    |x_{qn} - x_{rn}|
  \end{bmatrix}
\end{equation}

\noindent where $| \cdot |$ represents the absolute value of the difference, $x_{qi}$ and $x_{ri}$ are the $i$-th features of the signatures $\textbf{x}_q$ and $\textbf{x}_r$ respectively, and $n$ is the number of features. Hence, each dimension of the $\textbf{u}$ vector is equal to the distance between the corresponding dimensions of the vectors $\textbf{x}_q$ and $\textbf{x}_r$, and therefore all these vectors have the same dimensionality \citep{bertolini:16}.

As mentioned, regardless of the number of writers, after applying DT, only two classes are present in the dissimilarity space:
\begin{itemize}
    \item The \textit{within/positive class $w_+$}: the intraclass dissimilarity vectors, i.e., computed from samples of the same writer.
    \item The \textit{between/negative class $w_-$}: the interclass dissimilarity vectors, i.e., computed from samples of different writers.
\end{itemize}
Once the data is transposed into the dissimilarity space, a 2-class classifier (known as dichotomizer) is trained and used to perform the verification task. 
A common practice for WI systems is to use disjoint subsets of writers to train the classifier and to perform the verification. In general, the training set is known as the development set $D$ and the test set as exploitation set $\varepsilon$ \citep{cha:00}. 

The Dichotomy Transformation has already been used in various contexts, such as: bird species identification \cite{zottesso:18}, forest species recognition \cite{martins:15}, writer identification \cite{bertolini:16} and also for handwritten signature verification \cite{rivard:13,eskander:13,souza:18}.

Based on the DT definition we can already highlight the following points: ($C_1$) first of all, the DT reduces the high number of classes (writers) to a 2-class problem, and only one model is trained to perform the verification for all writers from the dissimilarity space (DS) generated by the dichotomy transformation \citep{eskander:13}.
($C_6$) The WI verification only depends on the reference signature used as input to the classifier; it means that the WI framework is scalable and can easily manage new incoming writers without requiring additional training or updating of the model (unlike the WD approach, where a new classifier needs to be trained).
In this way, a WI classifier trained in one dataset can be used to verify signatures from other datasets in a transfer learning task. In this scenario, the different datasets would represent samples that belong to the same domain (signature representations in DS). As defined before, given that the development set $D$ and the exploitation set $\varepsilon$ are disjoint, by default this approach already operates by using transfer learning.

An important property of the dichotomy transformation is its ability to increase the number of samples in the dissimilarity space since it is composed of each pairwise comparisons of signatures. Thus, if $M$ writers provide a set of $R$ reference signatures each, Equation \ref{eq:DT_distance} generates up to \begin{math}(^{M R}_{\;\;2})\end{math} different distances vectors. Of these, \begin{math}M(^{R}_{2})\end{math} are from the positive class and \begin{math}(^{M}_{2})R^2\end{math} belong to the negative class \citep{rivard:13}. Therefore, even with a small number of reference signatures per writer, DT can generate a large amount of samples in DS. 

In this way, the model can handle the small number of samples per class. Also, by increasing the number of samples, the model may be able to obtain sufficient information to capture the full range of signature variations, reducing the effects of the intra-class variability \citep{hafemann_review:17} ($C_3$).
Besides, by generating the same number of samples for both the positive class (questioned signatures are the genuine signatures from the writers) and the negative class (questioned signatures are the random forgeries), the model can manage the dataset imbalance ($C_4$).

However, many of the samples generated by DT in the WI-HSV scenario represent redundant information and therefore have little importance for training purposes. In \citep{souza_dissimilarity:19} and \citep{ souza_characterization:19} we showed that using prototype selection (PS) techniques, such as the Condensed Nearest Neighbors (CNN) \citep{hart:68}, in DS, allowed the WI classifier to be trained in the preprocessed DS without deteriorating the verification performance.

Another aspect is faced when writers have more than one reference signature. In this case, the pairwise comparison of DT is applied considering the questioned signature and each of the references, producing a set of dissimilarity vectors $\{{\textbf{u}_r}\}^{R}_{1}$, where $R$ is the number of reference signatures belonging to the writer. 
Thus, the dichotomizer evaluates each dissimilarity vector individually and produces a set of partial decisions $\{f(\textbf{u}_r)\}^{R}_{1}$ \citep{rivard:13}.
The final decision about the questioned signature is based on the fusion of all partial decisions by a function $g( \cdot )$ and depends on the output of the dichotomizer. For discrete output classifiers, the majority vote can be used; whereas for distance or probability outputs, the max, mean, median, min and sum functions may be applied \citep{rivard:13}.


It is important to mention that DT has already been used in the handwritten signature verification scenario \citep{rivard:13, eskander:13}, but using older feature representations. An important aspect of this transformation is the needing for a good feature representation, as the one used in this paper. The motivation for this statement is as follows: (i) signatures that are close in the feature space will be close to the origin in the dissimilarity space, this behavior is expected for genuine signatures. (ii) the further away two signatures are in the feature space, the farther the vector resulting from the dichotomy transformation will be from the origin. It is expected to find this behavior for the forgeries \citep{cha:00}. 
To complete the reasoning, as depicted in Figure \ref{fig:tsne}, this scenario can actually be found in the feature space from \textit{SigNet} \citep{hafemann:17}, as different writers are clustered in separate regions. This feature representation is discussed in section \ref{sec:feature_representation}.

Finally, one possible drawback of DT is that, perfectly grouped writers in the feature space may not be perfectly separated in the dissimilarity space \citep{cha:00}. Thus, the greater the dispersion between sample distributions among the writers, the less the dichotomizer is able to detect real differences between similar signatures \citep{rivard:13}.

Summarizing, based on the main properties of the WI dichotomy transformation, this approach can handle some data difficulties of the HSV problems that WD systems are not capable of. Other characteristics of DT can be found in \citep{cha:00, rivard:13}.

\begin{figure}[!htb]
\centering
  \includegraphics[width=\columnwidth]{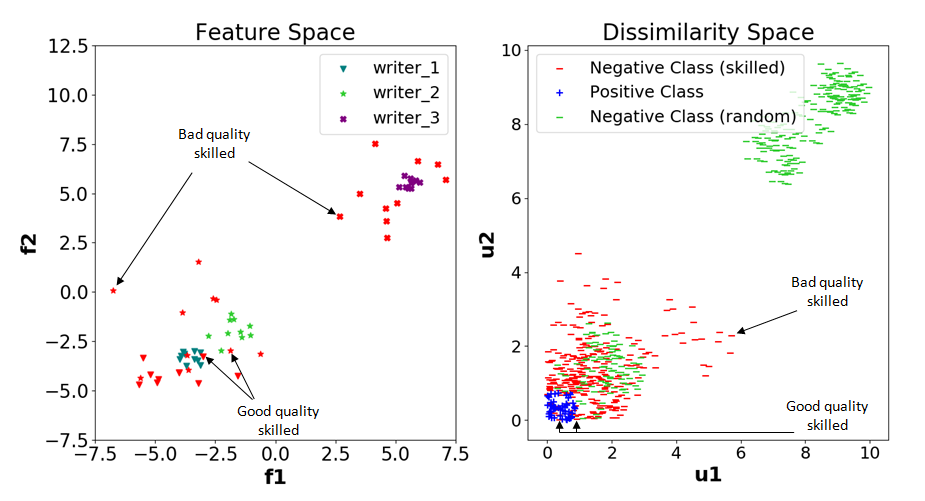}
  \caption{On the left, the feature space containing genuine signatures and skilled forgeries from 3 different writers. On the right, the dissimilarity space generated after applying the dichotomy transformation.}
  \label{fig:global_scenarios}
\end{figure}

To facilitate the understanding of DT, Figure \ref{fig:global_scenarios} (left) depicts a synthetic 2D feature space with synthetic data (containing genuine signatures and skilled forgeries from 3 different writers); on the right the respective dichotomy transformation is shown. The skilled forgeries in the feature space for each writer are presented in red with the same marker. These data were generated based on what was observed in Figure \ref{fig:tsne}. The reader should keep in mind that although the negative samples in the dissimilarity space are represented by different colors (red for the ones generated by the skilled forgeries and green for the random forgeries), they are part of the same class. This separation was made to support further discussions that will be held later.

Signatures that belong to the same writer are close to each other in the feature space. Hence, they will form a cluster located close to the origin in DS.
The quality of a forgery can be measured by its proximity to a target signature \citep{houmani:11}; this proximity should be considered in the feature space. When transposed to the DS, it is expected that, while bad quality skilled forgeries generate negative samples more distant to the origin, good quality skilled forgeries generate samples closer to the origin, and may even be within the positive cluster.

\subsection{Feature representation}
\label{sec:feature_representation}

The \textit{SigNet}, proposed by \cite{hafemann:17}, uses Deep Convolutional Neural Networks (DCNN) for learning the signature representations in a writer-independent way and, nowadays, represents a state of the art approach in this research area.
This approach tries to build a new representation space in which different writers are clustered in separate regions, based on the most representative properties of the handwritten signatures. To achieve this, the DCNN is trained by minimizing the negative log likelihood of the correct writer given the signature image. Table \ref{tab:dcnn_summary} summarizes the DCNN architecture used by the \textit{SigNet} model.

\begin{table}[!htb]
\caption{Summary of the \textit{SigNet} layers}
\label{tab:dcnn_summary}
\scriptsize
\centering

\begin{tabular}{lll}
\hline
Layer & Size & Other Parameters \\ 
\hline
Input & 1 x 150 x 220 & \\
Convolution (C1) & 96 x 11 x 11 & Stride = 4, pad = 0 \\
Pooling & 96 x 3 x 3 & Stride = 2 \\
Convolution (C2) & 256 x 5 x 5 & Stride = 1, pad = 2 \\
Pooling & 256 x 3 x 3 & Stride = 2 \\
Convolution (C3) & 384 x 3 x 3 & Stride = 1, pad = 1 \\
Convolution (C4) & 384 x 3 x 3 & Stride = 1, pad = 1 \\
Convolution (C5) & 256 x 3 x 3 & Stride = 1, pad = 1 \\
Pooling & 256 x 3 x 3 & Stride = 2 \\
Fully Connected (FC6) & 2048 & \\
\textbf{Fully Connected (FC7)} & \textbf{2048} & \\
Fully Connected + Softmax ($P(\textbf{y}|X)$) & M & \\
\hline

\end{tabular}
\end{table}

In the paper by \cite{hafemann:17}, the authors present another DCNN architecture, called as \textit{SigNet-f}, which uses skilled forgeries during the feature learning process. Our option of using the \textit{SigNet} is due to the fact that it is not reasonable to expect skilled forgeries to be available in the training phase for all users enrolled in the system.

For new writers, \textit{SigNet} is used to project the signature images onto the new representation space, by using feed-forward propagation until the FC7 layer, obtaining feature vectors with 2048 dimensions.
Also, as a writer-independent approach, it has the advantage of not being specific for a particular set of writers.

\begin{figure}[!htb]
\centering
  \includegraphics[width=\columnwidth]{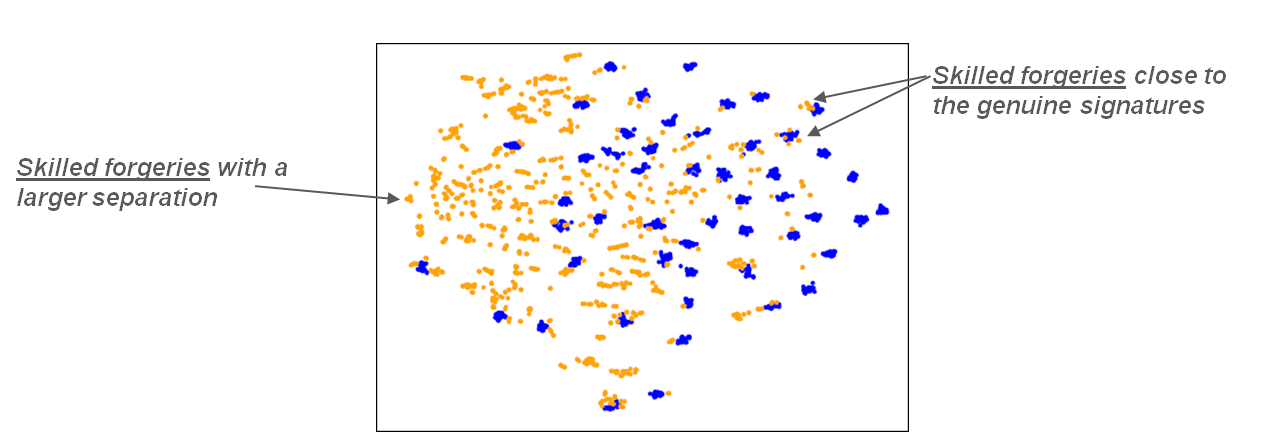}
  \caption{t-SNE 2D feature vector projections from the 50 writers of $V_v$. While blue points represent genuine signatures,  orange points represent skilled forgeries}
  \label{fig:tsne}
\end{figure}

In their study, \cite{hafemann:17} analysed the local structure of the learned feature space, by using the t-SNE algorithm in a subset containing 50 writers from the development set of the GPDS-300 dataset (referred to as the validation set for verification $V_v$). Figure \ref{fig:tsne} represents this analysis \citep{hafemann:17}.

As depicted in Figure \ref{fig:tsne}, in this feature space, for each writer, genuine signatures form compact clusters. 
According to \cite{houmani:11}, the forgery quality measures the proximity of a forgery to a target signature.
Thus, as highlighted, skilled forgeries come up with two different behaviors: (i) in some cases they have a larger separation from the genuine signatures. These forgeries are referred to as ``bad quality skilled forgeries'' in this paper. (ii) For some writers, the skilled forgeries are closer to the genuine signatures - we call them ``good quality skilled forgeries''. 

In this work, our original feature space is represented by the 2048 features obtained from the FC7 layer of the \textit{SigNet} \citep{hafemann:17} (available online\footnote{\raggedright\url{http://en.etsmtl.ca/Unites-de-recherche/LIVIA/Recherche-et-innovation/Projets/Signature-Verification}}). This model was chosen mainly because of its behavior, characterized by different writers clustered in separate regions of the feature space.

\subsection{Transfer Learning (TL)}

Transfer learning (TL) methods are based on the idea of utilizing the knowledge acquired from previously learned tasks, applying them to solve newer, related ones \citep{shao:15}. \cite{pan:10} present a formal definition for transfer learning. Given a source domain $D_S$ and a learning task $T_S$, a target domain $D_T$ and a learning task $T_T$, transfer learning aims to help improve the learning of the target predictive function $f_T (\cdot)$ in $D_T$ using the knowledge obtained from $D_S$ and $T_S$, where $D_S \neq D_T$, or $T_S \neq T_T$.

Following their notation, our context is related to the scenario where the target and the source domains are the same, i.e., $D_S = D_T$, and the learning tasks $T_S$ and $T_T$ are different. Specifically our case is that in which the conditional probability distributions of the domains are different, i.e., $P(Y_S|X_S) \neq P(Y_T|X_T)$, where $Y_{S_i}$ and $Y_{T_i}$ belong to the same label space formed by the positive and negative classes of the dichotomy transformation.

\cite{pan:10} suggest the following issues when dealing with transfer learning: 
\begin{itemize}
    \item What to transfer: the concern is related to which part of the knowledge may be common between the different domains, and so, may actually be useful to improve the performance in the target domain.
    \item How to transfer: methodologies need to be developed to deal with problems that may appear, such as the data distribution mismatch. Mining shared patterns from different domains, for instance, can significantly reduce the difference in the distribution between the target and the source domains
    \item When to transfer: considers in which situations TL should be used. When the domains are not related to each other, brute-force transfer may not succeed and/or even negatively affect the performance of learning in the target domain (situation knows as negative transfer).

\end{itemize}


\subsection{Instance Hardness (IH)}

Instance hardness (IH) measure is used to identify hard to classify samples \citep{smith:14}.
According to the paper by \cite{lorena:18}, an advantage of using the IH is to understand the difficulty of a problem at the instance level, rather than at the aggregated level with the entire dataset. 

For instance, in the paper by \cite{cruzKDN:17}, the authors used IH to identify the scenarios where an ensemble with dynamic selection techniques outperform the K-NN classifier.
The IH has also been used in ensemble generation methods \citep{walmsley:18, mariana_souza:18, kabir:18}.

The kDisagreeing Neighbors (kDN) measure is used herein to estimate IH. 
It represents the percentage of the $K$ nearest neighbors that do not share the label of a target instance.
This metric was chosen because it is able to capture the occurrence of class overlap and is also correlated with the frequency of a given instance being misclassified \citep{smith:14}. 
In a more formal definition, the kDN measure, $kDN(x_q)$, of a query instance $x_q$, whose K nearest neighbors are denoted by $KNN(x_q)$, is defined as:

\begin{equation}
\label{eq:kdn}
    kDN(x_q) = \frac{| x_k : x_k \in KNN(x_q) \wedge label(x_k) \neq label(x_q) |}{K}
\end{equation}

\noindent where $x_k$ represents a neighborhood instance and, $label(x_q)$ and $label(x_k)$ represent the class labels of the instances $x_q$ and $x_k$ respectively \citep{smith:14}. 

\begin{figure}[!htb]
\centering
  \includegraphics[width=\columnwidth]{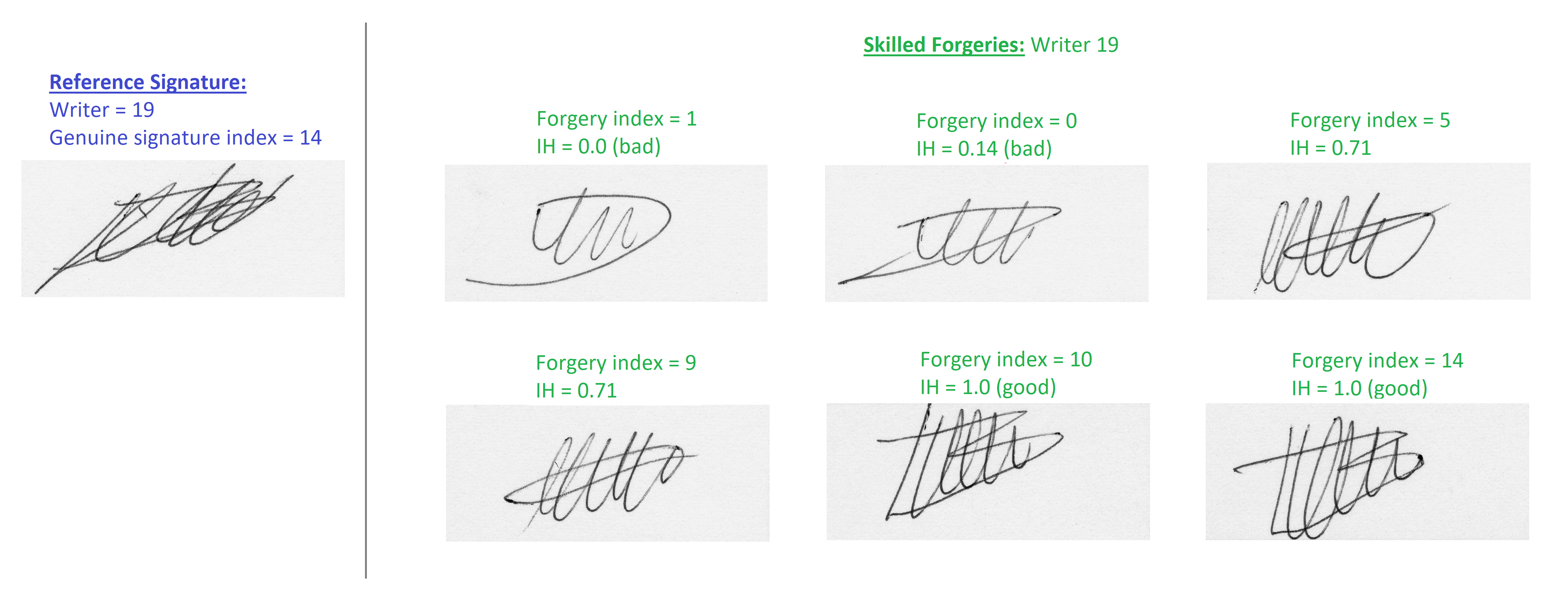}
  \caption{Good and bad skilled forgeries at image level}
  \label{fig:good_bad_skilled}
\end{figure}

Figure \ref{fig:good_bad_skilled} depicts examples of good and bad skilled forgeries at the image level, for the MCYT dataset. 
On the left, the genuine signature used as a reference is shown; the skilled forgeries are shown on the right.
It is expected that good quality skilled forgeries be more similar to the genuine signature than the bad ones. 
As previously presented, it is expected that the negative samples from the good skilled forgeries be close to the DS origin (as depicted in figure \ref{fig:global_scenarios}). Therefore, these negative samples may have more neighbors belonging to the positive class, i.e., higher IH values. 
On the other hand, as the negative samples from bad skilled forgeries present are more distant to the origin, they may have more neighbors belonging to the negative class, i.e., lower IH values. These aspects can also be seen in Figure \ref{fig:good_bad_skilled}.
Further discussions about these aspects are done in Section \ref{sec:results_discussion}.

\section{System overview}
\label{sec:overview}

\begin{figure}[!htb]
\centering
  \includegraphics[width=\columnwidth]{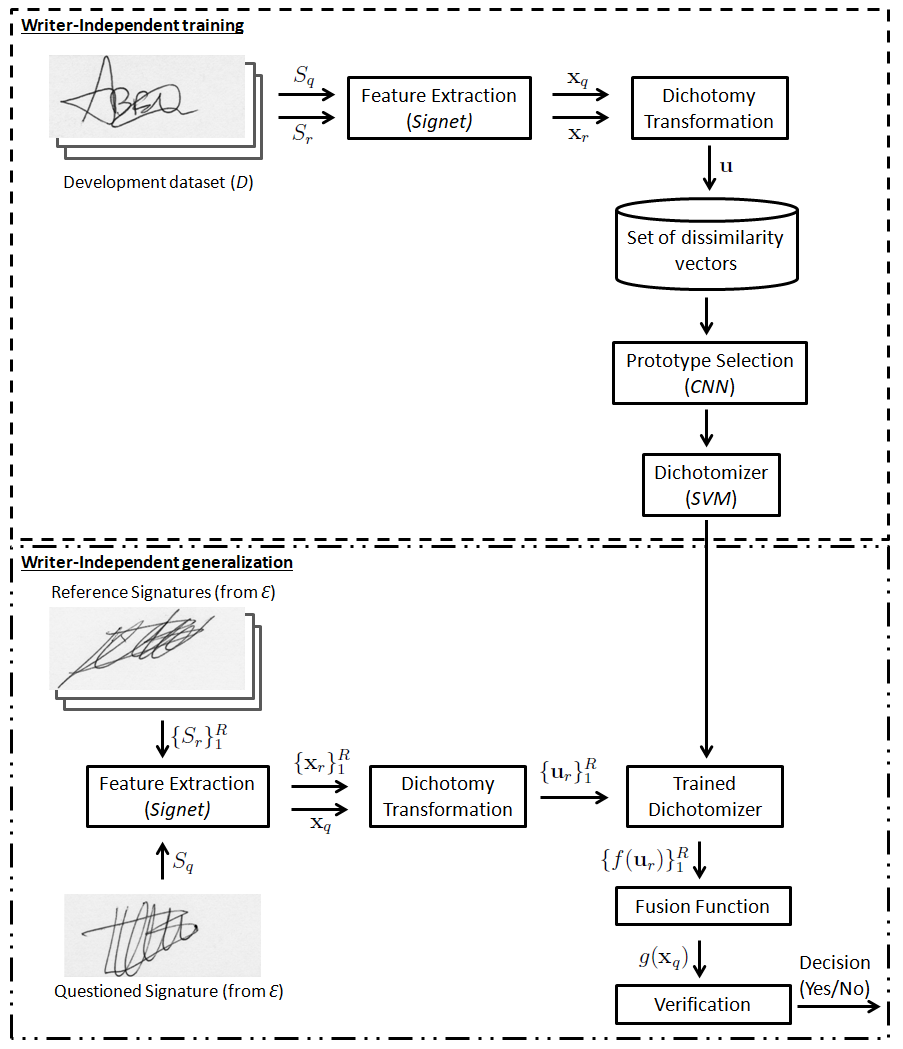}
  \caption{Block diagram containing the overview of the proposed approach.}
  \label{fig:block_diagram}
\end{figure}

Figure \ref{fig:block_diagram} depicts a block diagram containing the overview of the proposed approach. Top part of Figure \ref{fig:block_diagram} contains the training phase.
The first step is to obtain the feature vectors $\textbf{x}_q$ and $\textbf{x}_r$ extracted, respectively, from the images of the questioned signature $S_q$ and the reference signature $S_r$ belonging to the writers of the development dataset ($D$). This feature extraction is performed by using \textit{SigNet} (model is available online\footnote{\raggedright\url{https://github.com/luizgh/sigver_wiwd}}). Next, the dichotomy transformation is applied to obtain the dissimilarity vector $\textbf{u}$. After obtaining the set of dissimilarity vectors for all considered signatures of $D$, the Condensed Nearest Neighbors (CNN) is applied to discard redundant samples and then the dichotomizer is trained with the selected samples. The SVM was chosen as dichotomizer because it is considered one of the best classification methods for both WD and WI signature verification tasks \citep{hafemann_review:17}.

The generalization phase of the proposed approach is presented in the bottom part of Figure \ref{fig:block_diagram}.
Again, the first step is to extract the feature vectors through \textit{SigNet}. One difference from the training phase is that, here, a set of reference signatures $\{{S_r}\}^{R}_{1}$ is considered for each questioned signature $S_q$, both of them are obtained from the exploitation dataset ($\varepsilon$). Consequently a set of reference signatures, $\{{\textbf{x}_r}\}^{R}_{1}$, is considered in the dichotomy transformation. In this way, DT is applied considering the feature vector of the writer's questioned signature $\textbf{x}_q$ and the features vector set of his/her reference signatures $\{{\textbf{x}_r}\}^{R}_{1}$ and produces the set of dissimilarity vectors $\{{\textbf{u}_r}\}^{R}_{1}$. Next, the dichotomizer evaluates each dissimilarity vector individually and outputs a set of partial decisions $\{f(\textbf{u}_r)\}^{R}_{1}$. The final decision of the approach about the questioned signature is based on the fusion of all partial decisions by a function $g( \textbf{x}_q )$.

As the verification only depends on the input reference signature, by using the DT in a writer-independent approach, the dichotomizer can verify signatures of writers for whom the classifier was not trained (transfer learning). Thus, a single model already trained can be used to verify the signatures of new incoming writers, without any further transfer adaptation.

Our approach is centered on the dichotomy transformation. Thus, it presents the advantages and suffer the same weaknesses as this transformation (which were discussed and analysed in section \ref{sec:dichotomy_transformation}).

\section{Experiments}
\label{sec:experiments}

The objectives of the experiments are to analyse (i) the results obtained by the WI-classifier based on the dichotomy transformation, comparing them with the state-of-the-art, (ii) the dynamic reference selection through the MAX as fusion function, (iii) the accuracy of the model as a function of the IH in the GPDS and in transfer learning; and (iv) a characterization of good and bad quality skilled forgeries; 

\subsection{Datasets}
\label{sec:datasets}

The experiments are carried out using the GPDS, MCYT and CEDAR datasets, which are summarized in Table \ref{tab:datasets_summary}.

\begin{table}[!htb]
\caption{Summary of the used dataset.}
\label{tab:datasets_summary}
\scriptsize
\centering

\begin{tabular}{cccc}
\hline
Dataset Name & \#Writers & Genuine signatures (per writer) & Forgeries per writer \\ 
\hline
GPDS-960 & 881 & 24 & 30 \\ 
MCYT-75 & 75 & 15 & 15 \\ 
CEDAR & 55 & 24 & 24 \\ 
\hline

\end{tabular}
\end{table}

To enable comparison with other works, the GPDS-300 segmentation was used. 
In this case, the first 300 writers from the GPDS dataset form the exploitation set $\varepsilon$ and the development set $D$ is composed by the remaining 581 writers \citep{hafemann:17}.
It is worth noting that these subsets are disjoint, hence both of them are composed of different writers.

The training set is generated by using a subset of 14 genuine signatures for each writer from $D$ \citep{rivard:13, eskander:13}.
Samples belonging to the positive class are generated by applying the DT on the genuine signatures from every writer in $D$, as in Table \ref{tab:segmentation_gpds}. 
To have an equivalent number of counterexamples, the negative samples are generated by using 13 genuine signatures (references signatures) against one selected from a genuine signature of 7 different writers (7 random forgeries), as in Table \ref{tab:segmentation_gpds}. Thus, the same number of samples for both positive and negative classes are generated to be part of the training set.

\begin{table}[!htb]
\caption{GPDS-300 dataset: Development set $D$}
\label{tab:segmentation_gpds}
\scriptsize
\centering

\begin{tabular}{C{5cm}C{6cm}}
\hline
\multicolumn{2}{c}{Training set (14 signatures per writer)}  \\
\hline

Negative Class & Positive Class  \\ 
\hline
Pairwise comparisons among 13 signatures per writer and 7 random signatures of other writers & 
Pairwise comparisons among the 14 signatures per writer \\

\hline
$581 \cdot 13 \cdot 7 = 52,871$ negative samples &
$581 \cdot 14 \cdot 13 /2 = 52,871$ positive samples \\ 
\hline
\end{tabular}
\end{table}

In this study, the IH analyses are performed considering the neighborhood the GPDS-300 training set (after applying CNN preprocessing). So, to compute the IH value, each test sample is considered alone with the whole training set. Thus, we can observe the behavior of the test samples from different datasets in relation to the same training set neighborhood and, consequently, obtain a better understanding in a transfer learning scenario. The motivation for using the GPDS base training set is as follows: (i) GPDS has the largest training set, and (ii) as explained in section \ref{sec:feature_representation}, the features of the other datasets are obtained from a Deep Convolutional Neural Networks (DCNN) trained on GPDS dataset.


For CEDAR and MCYT datasets, in the scenarios where the classifiers are trained and tested in their own datasets the development sets are obtained as in \citep{souza_dissimilarity:19}. 
We used a 5x2 fold cross-validation. Using this methodology, at each time half of the writers are used for the development set while the other half for exploitation set.
As the CEDAR dataset has 55 writers, each fold would have 27 or 28 writers. 14 of the 24 genuine signatures of each writer in $D$ are randomly selected to generate the training set (Table \ref{tab:segmentation_cedar}).
In the MCYT dataset, the 75 writers were split into 37 or 38 writers per fold. From the 15 genuine signatures of each writer in $D$, 10 signatures are randomly selected to generate the training set (Table \ref{tab:segmentation_mcyt}). 
For the exploitation set in both of these datasets, the number of genuine signatures, skilled forgeries and random forgeries to be tested are the same as in Table \ref{tab:summary_exploitation_set}.
Segmentations are summarized, respectively, in Tables \ref{tab:segmentation_cedar} and \ref{tab:segmentation_mcyt}. In its turn, in the transfer learning scenario, the whole set of writers are used to obtain the development sets, but we keep the number of genuine signatures and random forgeries as in \citep{souza_dissimilarity:19}.


\begin{table}[!htb]
\caption{CEDAR dataset: Development set $D$}
\label{tab:segmentation_cedar}
\scriptsize
\centering

\begin{tabular}{C{5cm}C{6cm}}
\hline
\multicolumn{2}{c}{Training set (14 signatures per writer)}  \\
\hline

 Negative Class & Positive Class \\ 
\hline
Distances between the 13 signatures for each writer and 7 random signatures from other writers &
Distances between the 14 signatures for each writer ($D$)  \\ 
\hline
$(27 \; or \; 28) \cdot 13 \cdot 7$ samples & $(27 \; or \; 28) \cdot 14 \cdot 13 /2$ samples \\ 
\hline
\end{tabular}
\end{table}

\begin{table}[!htb]
\caption{MCYT dataset: Development set $D$ }
\label{tab:segmentation_mcyt}
\scriptsize
\centering

\begin{tabular}{C{5cm}C{6cm}}
\hline
\multicolumn{2}{c}{Training set (10 signatures per writer)}  \\
\hline

Negative Class & Positive Class \\ 
\hline
Distances between the 9 signatures for each writer and 5 random signatures from other writers & 
Distances between the 10 signatures for each writer ($D$)  \\ 

\hline
$(37 \; or \; 38) \cdot 9 \cdot 5$ samples & $(37 \; or \; 38) \cdot 10 \cdot 9 /2$ samples
\\ 
\hline
\end{tabular}
\end{table}

Considering that each dataset has a different number of writers and signature per writers and to be able to compare the results with the state-of-the-art the testing set is acquired using a methodology similar to that described in \citep{hafemann:17}. For the MCYT and the CEDAR datasets all the writers were used and differently from \citep{hafemann:17}, random forgeries were included. Table \ref{tab:summary_exploitation_set} summarizes the used Exploitation set $\varepsilon$ for each dataset.

\begin{table}[!htb]
\caption{Exploitation set $\varepsilon$}
\label{tab:summary_exploitation_set}
\scriptsize
\centering

\begin{tabular}{ccc}
\hline
Dataset & \#Samples & \#questioned signatures (per writer)\\ 
\hline
GPDS-300 & 9000 & 10 genuine, 10 skilled, 10 random  \\
MCYT & 2250 & 5 genuine, 15 skilled, 10 random  \\
CEDAR & 1650 & 10 genuine, 10 skilled, 10 random \\

\hline
\end{tabular}
\end{table}

\subsection{Experimental setup}

As first step, the distance vectors $\textbf{u}$ (in the dissimilarity space) are standardized (zero mean and unit variance). 
In the transfer learning scenarios, the same normalization from the training set is used for the other datasets (so the data is on the same scale).

Support Vector Machine (SVM) is considered one of the best classification methods for both WD and WI signature verification tasks \citep{hafemann_review:17}. 
In this paper, the SVM is used as writer-independent classifier with the following settings: $RBF$ kernel, $\gamma = 2^{-11}$ and $C=1.0$ ($C$ and $\gamma$ were selected based on a grid search: $C_{grid}$ = \{0.0001, 0.001, 0.01, 0.1, 1, 10, 100\} and $\gamma_{grid}$ = \{$2^{-11}$, 0.0001, 0.001, 0.01, 0.1, 1, 10, 100\}). The signed distance of the samples to the classifier's hyperplane are used as classifiers output \citep{hafemann:17, souza:18}. 

All the experiments in this study consider the training set after the Condensed Nearest Neighbors (CNN) preprocessing \citep{souza_dissimilarity:19, souza_characterization:19}. For the Condensed Nearest Neighbors (CNN), $K_{CNN} = 1$ \citep{hart:68}. For the instance hardness (IH) analysis, $K = 7$ is used for the estimation of the kDN \citep{cruzKDN:17}. For the comparative analysis, ten replications were carried out, as in \citep{souza_characterization:19}.

The Equal Error Rate ($EER$) metric, using user thresholds (considering just the genuine signatures and the skilled forgeries) was used in the evaluation of the verification models \citep{hafemann:17}. 
$EER$ is the error obtained when $FRR = FAR$, where (i) FRR (False Rejection Rate), represents the percentage of genuine signatures that are rejected by the system, and (ii) FAR (False Acceptance Rate), represents the percentage of forgeries that are accepted \citep{hafemann:17}. 

We have previously shown in \citep{souza:18} that, in general, by using MAX as fusion function and the highest number of references results in better performance. Thus, except when we analyse the fusion function, only this approach is considered in the scenario with multiple references.

\subsection{Results and discussion}
\label{sec:results_discussion}

\subsubsection{Comparison with the state of the art}

In this section we present the results on the GPDS-300 exploitation set, comparing the results with the state-of-the-art. 

\begin{table}[!htb]
\caption{Comparison of $EER$ with the state of the art in the GPDS-300 dataset (errors in \%)}
\label{tab:state_gpds}
\scriptsize
\centering

\begin{tabular}{ccccc}
\hline
Type & HSV Approach & \#Ref & \#Models & $EER$ \\ 
\hline
WI & \cite{kumar:12} & 1 & 1 &  13.76 \\ 
WI & \cite{eskander:13} & 1 & 1 &  17.82 \\ 
WI & \cite{dutta:16} & N/A & 1 &  11.21 \\ 
WI & \cite{hamadene:16} & 5 & 1 &  18.42 \\ 
WD & \cite{hafemann:16} & 12  &  300 & 12.83 \\ 
WD & \cite{soleimani:16} & 10  & 300 &  20.94 \\ 
WD & \cite{zois:16} & 5 & 300 &  5.48 \\ 
WD & \cite{hafemann:17} & 5 & 300 & 3.92 (0.18) \\ 
WD & \cite{hafemann:17} & 12 & 300 & 3.15 (0.18) \\ 
WD & \cite{serdouk:17} & 10 & 300 &  9.30 \\ 
WD & \cite{hafemann:18} & 12  & 300 & 3.15 (0.14) \\ 
WD & \cite{hafemann:18} (fine-tuned) & 12  & 300 & 0.41 (0.05) \\ 
WD & \cite{yilmaz:18} & 12  & 300 & 0.88 (0.36) \\ 
WD & \cite{zois:19} & 12 & 300 & 0.70 \\ 
WI & \cite{zois:19asymmetric} & 5 & 1 & 3.06 \\ 
\hdashline
WI & $SVM_{max}$ & 12 & 1 & 3.69 (0.18) \\ 
WI & $CNN\_SVM_{max}$ & 12 & 1 & 3.47 (0.15) \\ 
\hline
\end{tabular}
\end{table}

Table \ref{tab:state_gpds} contains both the comparison with the state of the art methods for the GPDS-300 dataset and also the results obtained by the WI-SVMs (with and without the CNN prototype selection). 

In general, our WI approach obtains low $EER$ that outperforms almost all other methods (eight of fourteen models), being comparable to \cite{hafemann:17} and \cite{hafemann:18}. It is overpassed only by the models reported in \cite{hafemann:18} (fine-tuned), \cite{yilmaz:18} and \cite{zois:19}. Although these models presented the best results, they are writer-dependent (WD); thus, our approach offers the advantage of being much more scalable, since only one classifier is used, while theirs requires 300.
Compared to the other WI models, our approach was able to outperform almost them all, except the model proposed by \cite{zois:19asymmetric}. 
It is worth noting that there is still room for improvement in our approach, such as, using ensemble or feature selection, which are approaches used in the paper by \cite{zois:19asymmetric}. 

\subsubsection{Fusion function}

In section \ref{sec:dichotomy_transformation}, we showed that when the writer has more than one reference signature, the dichotomizer produces a partial decision for each dissimilarity vector individually and merges them into the final decision.

\begin{figure}[!htb]
\centering
  \includegraphics[width=10cm]{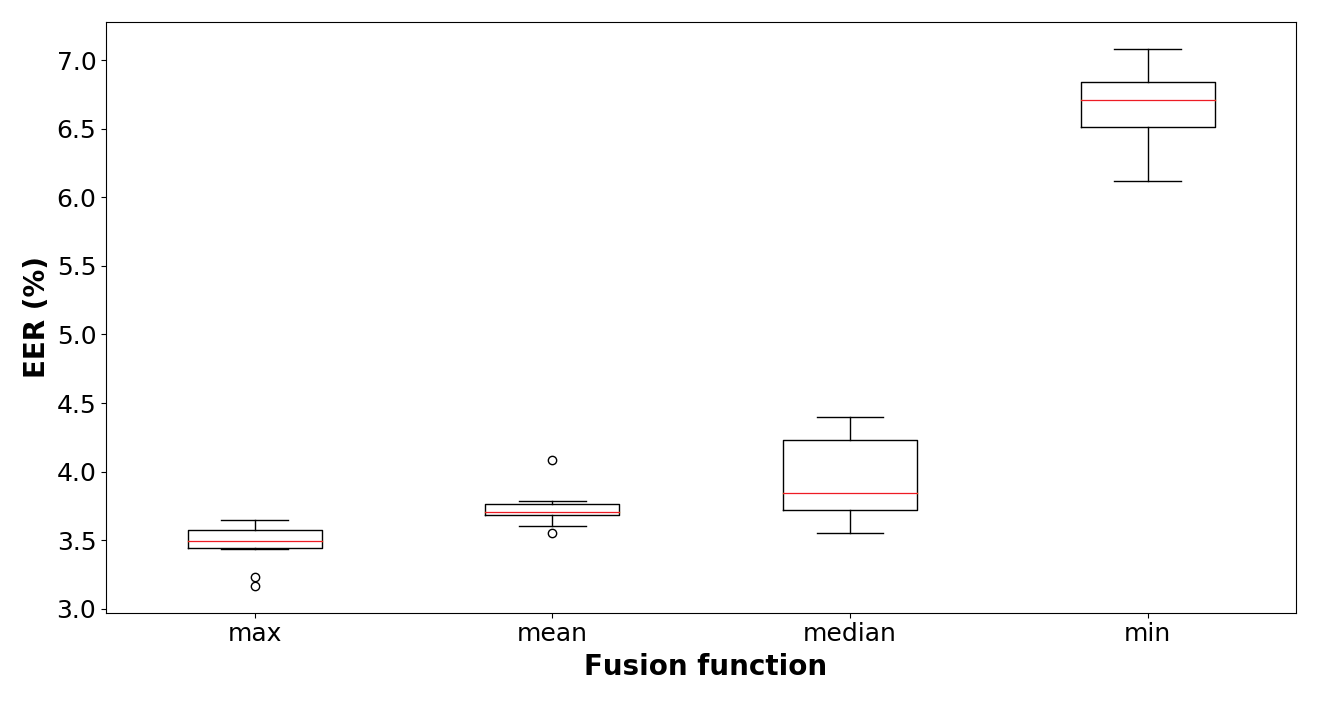}
  \caption{Boxplots for the EER (\%) on the GPDS-300 dataset, using 12 references. Given that it is a scenario with multiple references, the fusion functions MAX, MEAN, MEDIAN and MIN were considered.}
  \label{fig:boxplot}
\end{figure}

As we used the signed distance of a sample to the classifier’s hyperplane as classifiers output, the functions used in this fusion were: MAX, MEAN, MEDIAN, and MIN. As depicted in Figure \ref{fig:boxplot}, MAX obtained the best results considering ten replications with CNN-SVM model using 12 references for the GPDS-300 dataset.

\begin{figure}[!htb]
\centering
  \includegraphics[width=10cm]{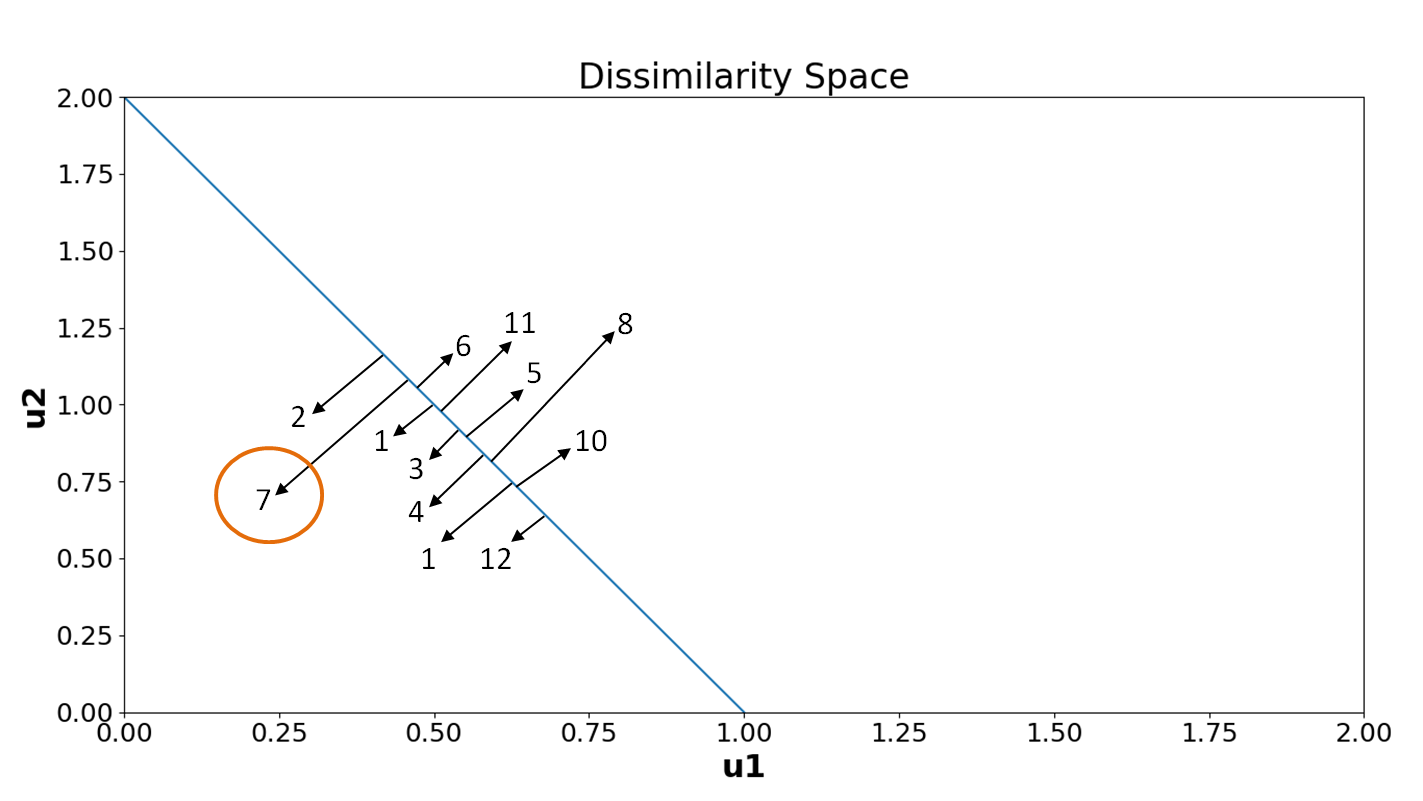}
  \caption{Dissimilarity space with the highlight on the selected reference, when MAX is used as a fusion function.}
  \label{fig:max_function}
\end{figure}

An important aspect related to the signed distance is that it indicates in which side of the hyperplane generated by the classifier the sample is located and how far it is from this hyperplane.
Figure \ref{fig:max_function} depicts this property. Given the dissimilarity space and the blue line representing a decision hyperplane with the left side as its positive side (because it is closer to the origin), then, for each fusion function, the distance used for the final decision would be:

\begin{itemize}
    \item MAX: the distance from the sample farthest from the hyperplane on the positive side;
    \item MIN: the distance from the sample farthest from the hyperplane on the negative side;
    \item MEAN and MEDIAN: respectively, the mean and median of all distances.
\end{itemize}

As we are in the dissimilarity space, the sample farthest from the hyperplane on the positive side represents the one that is closest to the DS origin, and, hence, the one generated by the DT of the reference signature and the questioned signature that are closest in the feature space. This happens when applying the MAX as fusion function and then, in the scenario of Figure \ref{fig:max_function}, sample 7 would be the one used to perform the verification.

On the other hand, the sample farthest from the hyperplane on the negative side represents the one that is further away from the DS origin, i.e., the one generated by the DT of the reference signature and the questioned signature that are farther apart from each other in the feature space. This represents the scenario of MIN as fusion function and therefore, in Figure \ref{fig:max_function}, sample 8 would be the one used to perform the verification.
In MEAN and MEDIAN, there is no specific sample selected since the mean and median of all distances is respectively used in each case.

Thus, when we apply the MAX as fusion function, the approach dynamically selects the sample closest to the origin in the dissimilarity space. Hence, it dynamically selects the reference (from the set of references) that is most similar to the questioned signature  and uses it to perform the verification.

\subsubsection{IH analysis}
\label{sec:ih_analysis}

In this section, some error evaluations considering the instance hardness are presented. For the skilled forgeries, some analysis considering the good and bad quality skilled forgeries were also carried out. 

Unlike previous works (\cite{souza_dissimilarity:19} and \cite{souza_characterization:19}), where the instance hardness was computed considering the test set.
In this study, the IH analyses are performed considering the neighborhood in the training set itself. Thus, to compute the IH value, each test sample is considered alone with the whole training set. Hence, in Equation \ref{eq:kdn}, the query instance, $x_q$, is a test sample and the K nearest neighbors, $KNN(x_q)$, belong to the training set.


We also extended the IH analyses to have a better understanding of the decision boundary (class overlap region). 
To this end, we present the relationship of IH values and the accuracy (\%) of the model when the user threshold of EER is used as the decision threshold.

Tables \ref{tab:reference_accuracy_IH_positive_training_nn}, \ref{tab:reference_accuracy_IH_negative_random_training_nn} and  \ref{tab:reference_accuracy_IH_negative_skilled_training_nn} present the relationship of the IH and the accuracy (\%) of the model when the user threshold of EER is used as decision threshold, respectively for the positive, negative (random) and negative (skilled) samples (for the GPDS dataset).
In the tables, the first column lists the IH values ($K = 7$) and the second column, the number of samples for the respective IH value.
The other columns represent the accuracy (\%) when considering the CNN-SVM and using, respectively, one, five, and twelve reference signatures.

First of all, we analyse the number of samples per IH value (second column).
As can be seen in Table \ref{tab:reference_accuracy_IH_positive_training_nn}, positive samples presented a major concentration in the $IH = 0.0$ bin and almost all of them had $IH <= 0.14$, which shows that the positive samples form a compact cluster. 

On the other hand, the negative samples were distributed along the IH values (Tables \ref{tab:reference_accuracy_IH_negative_random_training_nn} and  \ref{tab:reference_accuracy_IH_negative_skilled_training_nn}); this indicates that they are more sparsely distributed in the dissimilarity space than the positive samples. 
As the negative samples present higher IH values, including $IH = 1.0$, there may be an overlap of the classes, i.e., negative samples located inside the positive region of the dissimilarity space (all the negative sample neighbors belong to the positive class). These aspects are illustrated in the right part of figure \ref{fig:global_scenarios}.  

Moreover, the dashed line represents the limit where a kNN with $K = 7$ classifier could perform the correct classification.
As can be seen, a kNN classifier would obtain good results for the positive samples (due to the dense positive cluster), but it does not perform very well for the negative samples. For the negative samples, the high dimensionality, the data sparsity, the class overlap, and the presence of negative samples in the positive region of the dissimilarity space indicate the need for a strong discriminant classifier that can model complex distributions. That is why a kNN classifier fails on the classification. However, the CNN-SVM with \textit{RBF} kernel can deal with it and obtains better results even operating with one reference.

\begin{table}[!htb]
\caption{Relationship between IH and accuracy (\%) for the positive samples, for the GPDS-300 dataset}

\label{tab:reference_accuracy_IH_positive_training_nn}
\scriptsize
\centering

\begin{tabular}{ccccc}
\hline
IH & \#Samples & R1 & R5$_{max}$ & R12$_{max}$\\ 
\hline

\textbf{0.00} & \textbf{2330} & \textbf{95.02}  & \textbf{96.26}  &  \textbf{97.03} \\
0.14 & 591  & 90.18  & 94.07  &  94.75 \\
0.28 & 69   & 71.01  & 84.05  &  88.40 \\
0.42 & 6    & 66.66  & 83.33  &  100.00 \\
\hdashline
0.57 & 3    & 0.00   & 33.33  &  66.66  \\
0.71 & 1    & 100.00 & 100.00  & 100.00  \\
0.85 & 0    & - & - &  - \\
1.00 & 0    & - & - &  - \\
\hline
\end{tabular}
\end{table}

\begin{table}[!htb]
\caption{Relationship between IH and accuracy (\%) for the negative (random) samples, for the GPDS-300 dataset}
\label{tab:reference_accuracy_IH_negative_random_training_nn}
\scriptsize
\centering

\begin{tabular}{ccccc}
\hline
IH & \#Samples & R1 & R5$_{max}$ & R12$_{max}$\\ 
\hline

0.00 & 498 & 100.00 & 100.00 &  100.00  \\
0.14 & 488 & 100.00 & 100.00 &  100.00\\
0.28 & 461 & 100.00 & 100.00 &  100.00 \\
0.42 & 415 & 100.00 & 100.00 &  100.00 \\
\hdashline
0.57 & 418 & 100.00 & 100.00 &  100.00 \\
0.71 & 323 & 99.38 & 99.69 &  99.69 \\
0.85 & 276  & 99.27  & 100.00 &   100.00  \\
1.00 & 121 & 99.17  & 100.00 &   100.00 \\
\hline
\end{tabular}
\end{table}

\begin{table}[!htb]
\caption{Relationship between IH and accuracy (\%) for the negative (skilled) samples, for the GPDS-300 dataset}
\label{tab:reference_accuracy_IH_negative_skilled_training_nn}
\scriptsize
\centering

\begin{tabular}{ccccc}
\hline
IH & \#Samples & R1 & R5$_{max}$ & R12$_{max}$\\ 
\hline

0.00 & 420 & 100.00 & 100.00 &  100.00  \\
0.14 & 284 & 100.00 & 100.00 &  100.00\\
0.28 & 219 & 100.00 & 100.00  &  100.00 \\
0.42 & 208 & 100.00 & 100.00 &  99.51 \\
\hdashline
0.57 & 239 & 99.58  & 97.90 &  99.16 \\
0.71 & 348 & 95.86  & 97.70  &  97.98 \\
0.85 & 562 & 90.92  & 93.41  &  94.30 \\
\textbf{1.00} & \textbf{720} & \textbf{81.52}  & \textbf{88.05}  &  \textbf{90.69} \\
\hline
\end{tabular}
\end{table} 

The overlap in the positive region of the DS and the necessity of a more complex decision boundary can also be observed in the first row of Table \ref{tab:reference_accuracy_IH_positive_training_nn} and the last row of Table \ref{tab:reference_accuracy_IH_negative_skilled_training_nn} (as highlighted). 
From Table \ref{tab:reference_accuracy_IH_positive_training_nn}, first line, notice that all neighborhood instances from the positive samples belong to the positive class itself ($IH = 0.0$). Even so, the classifier did not achieve a perfect classification. 
In the same way, from Table \ref{tab:reference_accuracy_IH_negative_skilled_training_nn}, the classifier can correctly classify most of the negative (skilled) samples presenting the neighborhood formed by the positive class ($IH = 1.0$).

\begin{figure}[!htb]
\centering
  \includegraphics[width=8cm]{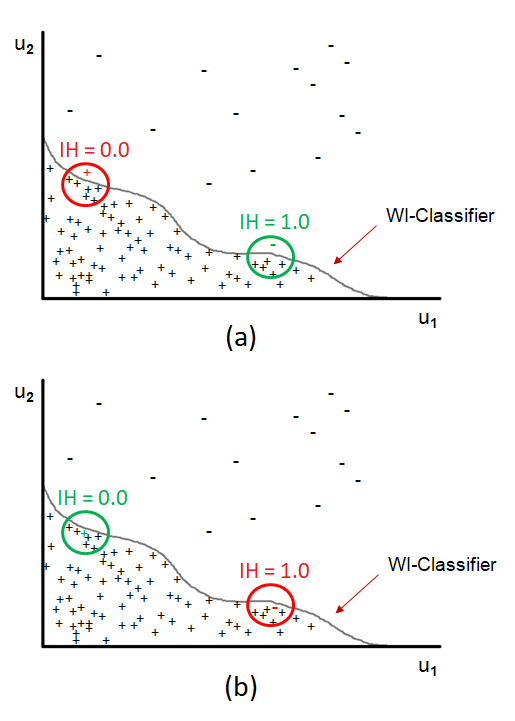}
  \caption{Synthetic decision frontiers: in (a) the scenario where the seven neighbors belong to the positive class and the model was able to correctly classify the sample from the negative class ($IH = 1.0$) but wrongly classified the positive test sample. In (b) the opposite scenario. }
  \label{fig:decision_frontier}
\end{figure}

Figure \ref{fig:decision_frontier} depicts this behavior in a synthetic representation.
Considering the presented WI-classifier decision frontier, in (a) in the cases where the seven neighbors belong to the positive class the model was able to correctly classify the sample from the negative class ($IH = 1.0$) but wrongly classified the positive test sample ($IH = 0.0$). Figure \ref{fig:decision_frontier} (b) illustrates the opposite scenario.

Specifically for the negative (skilled) forgeries, if we consider this same kNN limit to characterize the ``bad quality skilled forgeries'' ($IH <= 0.5$) and the ``good quality skilled forgeries'' ($IH > 0.5$). 
We can see that the CNN-SVM has an almost perfect performance for the bad quality skilled forgeries, independently of the number of references used (see the first four lines of Table \ref{tab:reference_accuracy_IH_negative_skilled_training_nn}). However, the higher the number of references used, the better the verification for the good quality skilled forgeries (the last four lines of Table \ref{tab:reference_accuracy_IH_negative_skilled_training_nn}).

From Table \ref{tab:reference_accuracy_IH_negative_random_training_nn}, the negative (random) samples are arranged along the IH values. This indicates that these samples are located in a sparse region of the dissimilarity space and some samples are closer to the region of the compact positive cluster in the space, because of the $IH = 1.0$ samples. However, the positive and the negative (random) sets may be disjoint, as the classifier presents an almost perfect verification performance.
These aspects can also be seen in the right part of figure \ref{fig:global_scenarios}.

\subsubsection{Transfer learning}

In \citep{souza_dissimilarity:19}, we experimentally showed that a WI-SVM trained in the GPDS can be employed to verify signatures in the other datasets without any further transfer adaptation in the WI-HSV context and still obtain similar results when compared to both WD and WI classifiers trained and tested in their own datasets.

In addition of using the CNN-SVM trained in the GPDS dataset (referred to as CNN-SVM$_{gpds}$ in this section), we also extend the transfer learning analysis by training WI-SVM models in CEDAR (CNN-SVM$_{cedar}$) and MCYT (CNN-SVM$_{mcyt}$) datasets and testing them in the other datasets. 
Table \ref{tab:transfer_learning_comparison} presents the comparison of these models. 


\begin{table}[!htb]
\caption{Comparison of $EER$ for both scenarios where models are trained and tested in their own datasets and transfer learning, for the considered datasets (errors in \%). 
CNN-SVM$_{cedar}$, CNN-SVM$_{mcyt}$ and CNN-SVM$_{gpds}$ are respectively the models trained in the CEDAR, MCYT and GPDS datasets. The models from \cite{zois:19asymmetric} follow the same terminology, so, $P_{2AD-cedar}$, $P_{2AD-mcyt}$, $P_{2AD-gpds}$ are respectively the models trained in the CEDAR, MCYT and GPDS datasets.}
\label{tab:transfer_learning_comparison}
\scriptsize
\centering
\begin{tabular}{ccccc}
\hline
Model  & \#Ref & $EER_{CEDAR}$ & $EER_{MCYT}$  & $EER_{GPDS-300}$ \\ 
\hline
\cite{zois:19asymmetric}: $P_{2AD-cedar}$ &  5  & 3.1  & 3.7 & 3.3 \\ 
\cite{zois:19asymmetric}: $P_{2AD-mcyt}$  &  5  & 2.9 &  4.6 & \textbf{2.9} \\ 
\cite{zois:19asymmetric}: $P_{2AD-gpds}$  &  5  & \textbf{2.8} &  3.4 & 3.7 \\
\hdashline
CNN-SVM$_{cedar}$      &  12  & 5.86 (0.50)  & 4.22 (0.77) & 5.42 (0.26)\\ 
CNN-SVM$_{mcyt}$       &  12  & 4.21 (0.37) &  2.99 (0.16) & 3.57 (0.10)\\ 
CNN-SVM$_{gpds}$       &  10  & 3.32 (0.22) &  \textbf{2.89 (0.13)} & 3.47 (0.15)\\
\hline
\end{tabular}
\end{table}

As can be seen in Table \ref{tab:transfer_learning_comparison}, for the proposed approach, transfer learning models were able to outperform the models trained and tested in the own datasets for the CEDAR dataset and obtained similar results for the MCYT and GPDS-300 datasets.
Which shows that the proposed approach can actually be used in a transfer learning context, reinforcing the scalability and adaptability of the WI systems.

In the paper by \cite{zois:19asymmetric}, the authors also used a transfer learning methodology. In this scenario, our approach obtains comparable results, being better in the MCYT dataset and worse when CEDAR and GPDS datasets are considered.

The state of the art results for CEDAR and MCYT datasets can be found, respectively, in Tables \ref{tab:state_cedar} and \ref{tab:state_mcyt}.
When compared to the state of the art models, despite having worse results than state of the art models, our approach was still able to obtain low verification error. However, compared to the other models that are mainly WD and, consequently, a classifier is required to be trained for each writer of the dataset, in our approach a single model is used to perform signature verification for all writers. 


Beyond the state of the art results, in Tables \ref{tab:state_cedar} and \ref{tab:state_mcyt} we also kept the results obtained by the CNN-SVM$_{gpds}$ model, since the IH analysis will be performed through it (the motivation of using this model was presented in section \ref{sec:datasets}).


\begin{table}[!htb]
\caption{Comparison of $EER$ with the state of the art in the CEDAR dataset (errors in \%)}
\label{tab:state_cedar}
\scriptsize
\centering
\begin{tabular}{ccccc}
\hline
Type & HSV Approach & \#Ref & \#Models & $EER$ \\ 
\hline
WI & \cite{kalera:04} & 16 & 1 & 21.9  \\ 
WI & \cite{chen:06} & 16 & 1 & 7.90  \\ 
WI & \cite{kumar:10} & 1 & 1 & 11.81 \\ 
WI & \cite{kumar:12} & 1 & 1 &  8.33 \\ 
WD & \cite{bharathi:13} & 12 & 55 & 7.84  \\ 
WD & \cite{ganapathi:13} & 14 & 55 & 6.01  \\ 
WD & \cite{shekar:13} & 16 & 55 & 9.58  \\ 
WI & \cite{kumar:14} & 16 & 1 & 6.02  \\ 
WI & \cite{guerbai:15} & 12 & 1 & 5.60  \\
WI & \cite{dutta:16} & N/A & 1 & 0.00  \\
WI & \cite{hamadene:16} & 5 & 1 & 2.11  \\
WD & \cite{okawa:16} & 16 & 55 & 1.60  \\ 
WD & \cite{serdouk:16} & 16 & 55 & 3.52  \\
WD & \cite{zois:16} & 5 & 55 & 4.12  \\ 
WD & \cite{hafemann:17} & 12 & 55 & 4.76 (0.36) \\ 
WD & \cite{zois:17} & 5 & 55 & 2.07  \\ 
WD & \cite{hafemann:18} & 10 & 55 & 3.60 (1.26)  \\ 
WD & \cite{hafemann:18} (fine-tuned) & 10 & 55 &  2.33 (0.88)\\ 
WD & \cite{okawa:18} & 16 & 55 & 1.00  \\ 
WD & \cite{tsourounis:18} & 5 & 55 & 2.82  \\
WD & \cite{zois:18} & 5 & 55 & 2.30  \\
WD & \cite{zois:19} & 10 & 55 & 0.79  \\
WI & \cite{zois:19asymmetric} & 5 & 1 & 2.90  \\
\hdashline
WI & CNN-SVM$_{gpds}$ & 12  & 1 & 3.32 (0.22) \\
\hline
\end{tabular}
\end{table}

\begin{table}[!htb]
\caption{Comparison of $EER$ with the state of the art in the MCYT dataset (errors in \%)}
\label{tab:state_mcyt}
\scriptsize
\centering
\begin{tabular}{ccccc}
\hline
Type & HSV Approach & \#Ref & \#Models & $EER$ \\ 
\hline
WD & \cite{fierrez:04} & 10 &  75 & 9.28 \\ 
WD & \cite{fernandez:07} & 5 & 75 & 22.4 \\ 
WD & \cite{gilperez:08} & 10 & 75 &  6.44 \\ 
WD & \cite{wen:09} & 5  & 75 &  15.02 \\ 
WD & \cite{vargas:11} & 10  & 75 &  7.08 \\ 
WD & \cite{ooi:16} & 10  & 75 &  9.87 \\ 
WD & \cite{soleimani:16} & 10  & 75 &  9.86 \\ 
WD & \cite{zois:16} & 5 & 75 & 6.02 \\ 
WD & \cite{hafemann:17} & 10 & 75  &  2.87 (0.42) \\ 
WD & \cite{serdouk:17} & 10 & 75 &  18.15\\ 
WD & \cite{zois:17} & 5 & 75 &  3.97\\ 
WD & \cite{hafemann:18} & 10 & 75 &  3.64 (1.04) \\ 
WD & \cite{hafemann:18} (fine-tuned) & 10 & 75 &  3.40 (1.08)\\ 
WD & \cite{okawa:18} & 10 & 75 &  6.40\\ 
WD & \cite{zois:18} & 5 & 75 &  3.52\\ 
WD & \cite{zois:19} & 10 & 75 &  1.37\\ 
WI & \cite{zois:19asymmetric} & 5 & 1 &  3.50\\ 
\hdashline
WI & CNN-SVM$_{gpds}$ & 10 & 1  & 2.89 (0.13) \\

\hline
\end{tabular}
\end{table}


From Tables \ref{tab:state_cedar} and \ref{tab:state_mcyt}, even operating in a transfer learning scenario our approach was able to obtain low verification errors that are at least comparable to the models derived from other state-of-the-art methods. 
When compared to the WD models, our approach outperforms seven from fourteen methods in CEDAR and is overpassed by only one of sixteen models in MCYT dataset. Still, our approach has the advantage of being scalable and using only one classifier to perform the verification. 
For the WI scenario, in the CEDAR dataset our approach presents better results than six of the nine models. When considering the MCYT dataset, our approach outperformed the results by \cite{zois:19asymmetric}.


It is worth noting that, when our WI-classifier is used in the transfer learning scenario, it never had access to data from other datasets different from the one in which he was trained. Thus, combining DT and the used features representation allowed the model to remove the bias from signature acquisition protocols of the different datasets (e.g., scanner, writing space, type of writing tool).


\paragraph{IH analysis in transfer learning}
Tables \ref{tab:mcyt_accuracy_IH_positive_training_nn}, \ref{tab:mcyt_accuracy_IH_negative_random_training_nn} and  \ref{tab:mcyt_accuracy_IH_negative_skilled_training_nn} present the relationship of IH and the accuracy (\%) of the model when the user threshold of EER is used as decision threshold, respectively for the positive, negative (random) and negative (skilled) samples, for the MCYT dataset.
In these tables, the first column represents the possible IH values ($K = 7$), in the second column, the number of samples for the respective IH value.
The other columns present the accuracy (\%) when using one, five and ten references to perform the verification.

\begin{table}[!htb]
\caption{Relationship between IH and accuracy (\%) for the positive samples, for the MCYT dataset}
\label{tab:mcyt_accuracy_IH_positive_training_nn}
\scriptsize
\centering

\begin{tabular}{ccccc}
\hline
IH & \#Samples & R1 & R5$_{max}$ & R10$_{max}$\\ 
\hline
\textbf{0.00} & \textbf{357}  & \textbf{94.95} & \textbf{97.19}  &  \textbf{97.75} \\
0.14 & 16   & 62.50 & 100.00 &  100.00 \\
0.28 & 1    & 0.00  & 100.00 &  100.00 \\
0.42 & 0    & - & - & -   \\
\hdashline
0.57 & 0    & - & - & -   \\
0.71 & 1    & 0.00  & 100.00 &  100.00 \\
0.85 & 0    & - & - & -   \\
1.00 & 0    & - & - & -  \\
\hline
\end{tabular}
\end{table}

\begin{table}[!htb]
\caption{Relationship between IH and accuracy (\%) for the negative samples (random), for the MCYT dataset }
\label{tab:mcyt_accuracy_IH_negative_random_training_nn}
\scriptsize
\centering
\begin{tabular}{ccccc}
\hline
IH & \#Samples & R1 & R5$_{max}$ & R10$_{max}$\\ 
\hline
0.00 & 9   & 100.00 &  100.00  & 100.00   \\
0.14 & 51  & 100.00 &  100.00  & 100.00    \\
0.28 & 63  & 100.00 &  100.00  & 100.00 \\
0.42 & 94  & 100.00 &  100.00  & 100.00  \\
\hdashline
0.57 & 109 & 100.00 &  100.00  & 100.00  \\
0.71 & 123 & 100.00 &  100.00  & 100.00  \\
0.85 & 160 & 99.37  &  100.00  & 100.00  \\
1.00 & 141 & 100.00 &  100.00  & 100.00  \\
\hline
\end{tabular}
\end{table}

\begin{table}[!htb]
\caption{Relationship between IH and accuracy (\%) for the negative (skilled) samples, for the MCYT dataset }
\label{tab:mcyt_accuracy_IH_negative_skilled_training_nn}
\scriptsize
\centering

\begin{tabular}{ccccc}
\hline
IH & \#Samples & R1 & R5$_{max}$ & R10$_{max}$\\ 
\hline
0.00 & 0    & - & - & -  \\
0.14 & 2    & 100.00 & 100.00 & 100.00  \\
0.28 & 9    & 100.00 & 100.00 & 100.00  \\
0.42 & 22   & 100.00 & 100.00 & 100.00  \\
\hdashline
0.57 & 34   & 97.05  & 100.00 & 100.00  \\
0.71 & 101  & 99.00  & 99.00 & 99.00  \\
0.85 & 255  & 96.47  & 98.43 & 98.43  \\
\textbf{1.00} & \textbf{702}  & \textbf{88.31}  & \textbf{95.86} & \textbf{96.29}  \\
\hline
\end{tabular}
\end{table} 

As discussed before, we can consider that the dissimilarity space from different datasets as samples that belong to the same domain (signature representations in DS).
Even the data here presenting different concentration of samples per IH value, the used CNN-SVM presented similar behavior in the error analysis, when compared to the GDPS-300 scenario.
From the first row of Table \ref{tab:mcyt_accuracy_IH_positive_training_nn},
the classifier did not achieve a perfect classification, even the positive samples presenting all their neighbor from the positive class ($IH = 0.0$).
From the last row of Table \ref{tab:mcyt_accuracy_IH_negative_skilled_training_nn}, the classifier was able to correctly classify most of the negative (skilled), even the samples presenting the neighborhood formed by the positive class ($IH = 1.0$).
Thus, this confirms the overlap in the positive region of the DS and the need for a more complex decision boundary.

From Table \ref{tab:mcyt_accuracy_IH_negative_random_training_nn}, the negative (random) samples are arranged along the IH values. This indicates the sparsity of these data in the dissimilarity space and that some samples are located closer to the compact positive region of the space, because of the samples with $IH = 1.0$. However, there is no overlap in the positive region, as the model achieved a perfect verification performance.

As can be seen in \ref{appendix:cedar}, the WI approach also presented similar behavior for the CEDAR datasets. 
Thus, in all the scenarios, positive samples form a dense cluster (almost all positive samples have $IH \leq 0.14$), and the negative samples are scattered throughout space. The negative (random) samples may be disjoint to the positive set. The negative samples formed by the ``good quality skilled forgeries'' overlap the positive region of the DS, resulting in the need for a classifier with complex decision boundary.

\section{Lessons learned}
\label{sec:lessons_learned}

In the handwritten signature verification problem, the WI framework based on the dichotomy transformation (DT) is scalable, adaptable and presents the benefit of being able to handle some of the challenges faced when dealing with the HSV problem. Among them,
($C_1$) the high number of writers (classes), ($C_3$) the small number of training samples per writer with high intra-class variability and ($C_4$) the heavily imbalanced class distributions. 

Another advantage of the WI framework is that it can easily manage new incoming writers ($C_6$), and may even be used in a transfer learning context since the different datasets would represent samples that belong to the same domain (signature representations in the dissimilarity space). However with different acquisition protocol (scanner, writing space, writing tool etc). Therefore, a single model already trained can be used to verify the signatures of new incoming writers without any further transfer adaptation.

However, having a good feature representation of the signatures, like the one used in this study (characterized by different writers clustered in separate regions of the feature space) is very important for DT. 
The greater the separation between writers in the feature space, the smaller the overlap between the positive and negative classes in the dissimilarity space.

Finally, based on the IH analysis, the overlap between positive and negative (skilled) samples is still present, so feature selection could be applied in the dissimilarity space in the attempt to separate these sets of samples.

\section{Conclusion}
\label{sec:conclusion}

In this study, we addressed the understanding of the DT applied in a WI framework for handwritten signatures verification in a white-box manner. The experimental evaluations, carried out in four datasets, were based on both the EER and IH metric, which allowed us to understand the difficulty of the HSV problem at the instance level. 

The reported IH analysis showed that the samples belonging to the positive class form a compact cluster located close to the origin and the negative samples are sparsely distributed in the dissimilarity space generated by the dichotomy transformation. 
Furthermore, we were able to characterize the good and bad quality skilled forgeries using the IH analysis and also the frontier between the hard to classify samples, which are genuine signatures and good skilled forgeries close to the frontier. 

The DT characteristics and the analyses reported in this paper serve as motivation for future works aiming at improving the discrimination between genuine signatures and forgeries, focusing mainly on discriminating between good quality skilled forgeries. Some suggestions for future work include: 
(i) feature selection 
and (ii) ensemble learning and dynamic selection adapted to work on the Dissimilarity Space.

\section*{Acknowledgment}

This work was supported by the CNPq (Conselho Nacional de Desenvolvimento Científico e Tecnológico), FACEPE (Fundação de Amparo à Ciência e Tecnologia de Pernambuco) and the École de Technologie Supérieure (ÉTS Montréal). 

\bibliography{mybibfile}

\appendix

\newpage
\section{CEDAR results}
\label{appendix:cedar}

Tables \ref{tab:cedar_accuracy_IH_positive_training_nn}, \ref{tab:cedar_accuracy_IH_negative_random_training_nn} and  \ref{tab:cedar_accuracy_IH_negative_skilled_training_nn} present the relationship of IH and the accuracy (\%) of the model when the user threshold of EER is used as decision threshold, respectively for the positive, negative (random) and negative (skilled) samples, for the CEDAR dataset.
In the tables, the first column represents the possible IH values ($K = 7$), in the second column the number of samples for the respective IH value.
The other columns represent the accuracy (\%) when considering the CNN-SVM trained in the training set of GPDS-300 dataset and using respectively one, five and twelve references.

\begin{table}[!htb]
\caption{Relationship between IH and accuracy (\%) for the positive samples, for the CEDAR dataset}
\label{tab:cedar_accuracy_IH_positive_training_nn}
\scriptsize
\centering

\begin{tabular}{ccccc}
\hline
IH & \#Samples & R1 & R5$_{max}$ & R12$_{max}$\\ 
\hline

\textbf{0.00} & \textbf{482} & \textbf{92.53} & \textbf{94.60} & \textbf{95.85} \\
0.14 & 60  & 80.00 & 93.33 & 95.00 \\
0.28 & 1   & 0.00 & 0.00 &  100.00\\
0.42 & 0   & - & - & - \\
\hdashline
0.57 & 0   & - & - & - \\
0.71 & 2   & 50.00 & 100.00 & 100.00  \\
0.85 & 1   & 0.00 & 100.00 &  100.00 \\
1.00 & 4   & 0.00 & 100.00 &  100.00 \\
\hline
\end{tabular}
\end{table}

\begin{table}[!htb]
\caption{Relationship between IH and accuracy (\%) for the negative (random) samples, for the CEDAR dataset }
\label{tab:cedar_accuracy_IH_negative_random_training_nn}
\scriptsize
\centering

\begin{tabular}{cccccc}
\hline
IH & \#Samples & R1 & R5$_{max}$  & R12$_{max}$\\ 
\hline

0.00 & 11  & 100.00 & 100.00 & 100.00 \\
0.14 & 68  & 95.58  & 100.00 & 100.00 \\
0.28 & 89  & 95.50  & 100.00 & 100.00 \\
0.42 & 58  & 98.27  & 100.00 & 100.00 \\
\hdashline
0.57 & 69  & 100.00 & 100.00 & 100.00 \\
0.71 & 79  & 100.00 & 100.00 & 100.00 \\
0.85 & 110 & 100.00 & 100.00 & 100.00 \\
1.00 & 66  & 100.00 & 100.00 & 100.00 \\
\hline
\end{tabular}
\end{table}

\begin{table}[!htb]
\caption{Relationship between IH and accuracy (\%) for the negative (skilled) samples, for the CEDAR dataset }
\label{tab:cedar_accuracy_IH_negative_skilled_training_nn}
\scriptsize
\centering

\begin{tabular}{ccccc}
\hline
IH & \#Samples & R1 & R5$_{max}$ & R12$_{max}$\\ 
\hline

0.00 & 1    & 100.00 & 100.00 & 100.00 \\
0.14 & 15   & 80.00  & 100.00 & 100.00 \\
0.28 & 16   & 100.00 & 100.00 & 100.00 \\
0.42 & 18   & 100.00 & 100.00 & 100.00 \\
\hdashline
0.57 & 29   & 100.00 & 100.00 & 100.00 \\
0.71 & 35   & 97.14  & 97.14  & 100.00 \\
    0.85 & 147  & 96.59  & 95.91  & 97.95 \\
\textbf{1.00} &  \textbf{289} & \textbf{87.54}  & \textbf{93.42}  & \textbf{94.46} \\

\hline
\end{tabular}
\end{table} 


\clearpage
\centerline{\textbf{\Large SUPPLEMENTAL MATERIAL}}

\vspace{5mm}




\section*{S1 Experiments on BRAZILIAN dataset}

The same set of experiments as in the original article is carried out using the BRAZILIAN dataset \cite{freitas:00}, which is summarized in Table \ref{tab:brazilian_summary}.


\begin{table}[!htb]
\caption{Summary of the BRAZILIAN dataset.}
\label{tab:brazilian_summary}
\scriptsize
\centering

\begin{tabular}{cccc}
\hline
Dataset Name & \#Writers & Genuine signatures (per writer) & Forgeries per writer \\ 
\hline
Brazilian (PUC-PR) & 60+108 & 40 & 10 Simple, 10 Skilled \\ 
\hline

\end{tabular}
\end{table}

For the BRAZILIAN dataset, in the scenarios where the WI-classifier is trained and tested in its own dataset, the development set is obtained as in \citep{souza_dissimilarity:19}, which is summarized in Table \ref{tab:segmentation_brazilian}. On the other hand, in the transfer learning scenario, the whole set of writers are used to obtain the development set, but we keep the number of genuine signatures and random forgeries as in \citep{souza_dissimilarity:19}.

\begin{table}[!htb]
\caption{BRAZILIAN dataset: Development set $D$}
\label{tab:segmentation_brazilian}
\scriptsize
\centering

\begin{tabular}{C{5cm}C{6cm}}
\hline
\multicolumn{2}{c}{Training set}  \\
\hline

Positive Class & Negative Class \\ 
\hline
Distances between the 30 signatures for each writer & 
Distances between the 29 signatures for each writer and 15 random signatures from other writers \\ 
\hline
$108 \cdot 30 \cdot 29 /2 = 46,980$ samples & $108\cdot 29 \cdot 15  = 46,980$ samples
\\ 
\hline
\end{tabular}
\end{table}

The Exploitation set $\varepsilon$ is acquired using a methodology similar to that described in \citep{hafemann:17}. In this dataset, only the first 60 writers have simple and skilled forgeries, so only these writers are considered during the test. 
Table \ref{tab:summary_brazilian_exploitation_set} summarizes the Exploitation set for the BRAZILIAN dataset. \\

\begin{table}[!htb]
\caption{Exploitation set $\varepsilon$}
\label{tab:summary_brazilian_exploitation_set}
\scriptsize
\centering

\begin{tabular}{ccc}
\hline
Dataset & \#Samples & \#questioned signatures (per writer)\\ 
\hline
BRAZILIAN & 2400 &  10 genuine, 10 random, 10 simple, 10 skilled   \\

\hline
\end{tabular}
\end{table}

\subsection*{S1.1 BRAZILIAN results}

In addition of using the CNN-SVM trained in the GPDS dataset (referred to as CNN-SVM$_{gpds}$), CEDAR (CNN-SVM$_{cedar}$) and MCYT (CNN-SVM$_{mcyt}$), we also extend the transfer learning analysis by training WI-SVM model in BRAZILIAN (CNN-SVM$_{brazilian}$) dataset and testing it in the other datasets. 
Also, testing all the models in the BRAZILIAN dataset.
Table \ref{tab:brazilian_transfer_learning_comparison} presents the comparison of these models. In it, while rows contains the models, columns contains the $EER$ for the datasets.


\begin{table}[!htb]
\caption{Comparison of $EER$ for both scenarios where models are trained and tested in their own datasets and transfer learning, for the considered datasets (errors in \%)}
\label{tab:brazilian_transfer_learning_comparison}
\scriptsize
\centering
\begin{tabular}{cccccc}
\hline
Model  & \#Ref & $EER_{BRAZILIAN}$ & $EER_{CEDAR}$ & $EER_{MCYT}$  & $EER_{GPDS-300}$ \\ 
\hline
CNN-SVM$_{brazilian}$   &  30   & 1.26 (0.33)  & \textbf{3.12 (0.41)} &  6.57 (0.33) & 7.35 (0.34) \\ 
CNN-SVM$_{cedar}$       &  12   & \textbf{0.72 (0.14)}  & 5.86 (0.50)  & 4.22 (0.77) & 5.42 (0.26)\\ 
CNN-SVM$_{mcyt}$       &  12   & 1.16 (0.29)  & 4.21 (0.37) &  2.99 (0.16) & 3.57 (0.10)\\ 
CNN-SVM$_{gpds}$       &  10   & 1.11 (0.37)  & 3.32 (0.22) &  \textbf{2.89 (0.13)} & \textbf{3.47 (0.15)}\\
\hline
\end{tabular}
\end{table}

As can be seen in Table \ref{tab:brazilian_transfer_learning_comparison}, for the proposed approach, when the verification is performed in BRAZILIAN dataset, transfer learning models were able to outperform the model trained and tested in the own dataset.
It is also worth noting that, the CNN-SVM$_{brazilian}$ model was the one that got the lowest $EER$ for the CEDAR database.
Both of these two aspects show that the proposed approach can actually be used in a transfer learning context, reinforcing the scalability and adaptability of the WI systems.


Table \ref{tab:state_brazilian} presents the state of the art results for the BRAZILIAN dataset. Beyond the state of the art results and the results from the CNN-SVM$_{brazilian}$, we also kept the results obtained by the CNN-SVM$_{gpds}$ model, since the IH analysis will be performed through it.


\begin{table}[!htb]
\caption{Comparison of $EER$ with the state of the art in the BRAZILIAN dataset (errors in \%)}
\label{tab:state_brazilian}
\scriptsize
\centering

\begin{tabular}{ccccc}
\hline
Type & HSV Approach & \#Ref & \#Models & $EER$ \\ 
\hline
WD & \cite{hafemann:16} & 15  & 60 &  4.17 \\ 
WD & \cite{hafemann:17} & 5  & 60 &  2.92 (0.44) \\ 
WD & \cite{hafemann:17} & 15  & 60 &  2.07 (0.63) \\ 
WD & \cite{hafemann:17} & 30  & 60 &  2.01 (0.43) \\ 
WD & \cite{hafemann:18} & 15  & 60 &  1.33 (0.65) \\ 
WD & \cite{hafemann:18} (finetuned) & 15  & 60 &  1.35 (0.60) \\ \hdashline

WI & CNN-SVM$_{brazilian}$ & 30 & 1 & 1.26 (0.33) \\
WI & CNN-SVM$_{gpds}$ & 30 & 1 & 1.11 (0.37) \\

\hline

\end{tabular}
\end{table}

From Tables \ref{tab:state_brazilian}, the CNN-SVM$_{brazilian}$, using a single model to perform signature verification for all writers, was able to obtain better results when compared to the WD models found in the literature (which use one classifier for each writer). Most interesting finding is that, even operating in a transfer learning scenario the CNN-SVM$_{gpds}$ model was able to outperform all methods for the BRAZILIAN dataset.

\paragraph{\textbf{IH analysis in transfer learning}}

Tables \ref{tab:brazilian_accuracy_IH_positive_training_nn}, \ref{tab:brazilian_accuracy_IH_negative_random_training_nn},  \ref{tab:brazilian_accuracy_IH_negative_simple_training_nn} and \ref{tab:brazilian_accuracy_IH_negative_skilled_training_nn} present the relationship of IH and the accuracy (\%) of the model when the user threshold of EER is used as decision threshold, respectively for the positive, negative (random), negative (simple) samples and negative (skilled) samples, for the BRAZILIAN dataset.
In the tables, the first column represents the possible IH values ($K = 7$), in the second column the number of samples for the respective IH value.
The other columns represent the accuracy (\%) when considering the CNN-SVM trained in the training set of GPDS-300 dataset and using respectively one, five, fifteen and thirty references.

\begin{table}[!htb]
\caption{Relationship between IH and accuracy (\%) for the positive samples, for the BRAZILIAN dataset}
\label{tab:brazilian_accuracy_IH_positive_training_nn}
\scriptsize
\centering

\begin{tabular}{cccccc}
\hline
IH & \#Samples & R1 & R5$_{max}$ & R15$_{max}$ & R30$_{max}$\\ 
\hline

\textbf{0.00} & \textbf{591} &  \textbf{96.27}  &  \textbf{97.63} &  \textbf{99.49}  & \textbf{99.66} \\
0.14 & 7   &  57.14  &  71.42 & 71.42   & 85.71 \\
0.28 & 1   &  100.00 &  100.00 &  100.00  & 100.00 \\
0.42 & 0   &  - &  - &  -  & - \\
\hdashline
0.57 & 1   &  0.00   & 0.00  &  100.00  &  100.00 \\
0.71 & 0   &  - &  - &  -  & - \\
0.85 & 0   &  - &  - &  -  & - \\
1.00 & 0   &  - &  - &  -  & - \\
\hline
\end{tabular}
\end{table}

\begin{table}[!htb]
\caption{Relationship between IH and accuracy (\%) for the negative (random) samples, for the BRAZILIAN dataset}
\label{tab:brazilian_accuracy_IH_negative_random_training_nn}
\scriptsize
\centering

\begin{tabular}{cccccc}
\hline
IH & \#Samples & R1 & R5$_{max}$ & R15$_{max}$ & R30$_{max}$\\ 
\hline

0.00 & 79  & 100.00  & 100.00  &  100.00  & 100.00 \\
0.14 & 122 & 100.00  & 100.00  &  100.00  & 100.00 \\
0.28 & 96  & 100.00  & 100.00  &  100.00  & 100.00 \\
0.42 & 72  & 100.00  & 100.00  &  100.00  & 100.00 \\
\hdashline
0.57 & 55  & 100.00  & 100.00  &  100.00  & 100.00 \\
0.71 & 47  & 100.00  & 100.00  &  100.00  & 100.00 \\
0.85 & 52  & 100.00  & 100.00  &  100.00  & 100.00 \\
1.00 & 77  & 100.00  & 100.00  &  100.00  & 100.00 \\
\hline
\end{tabular}
\end{table}

\begin{table}[!htb]
\caption{Relationship between IH and accuracy (\%) for the negative (simple) samples, for the BRAZILIAN dataset }
\label{tab:brazilian_accuracy_IH_negative_simple_training_nn}
\scriptsize
\centering

\begin{tabular}{cccccc}
\hline
IH & \#Samples & R1 & R5$_{max}$ & R15$_{max}$ & R30$_{max}$\\ 
\hline

0.00 & 69  & 100.00  & 100.00  &  100.00  & 100.00 \\
0.14 & 98  & 100.00  & 100.00  &  100.00  & 100.00 \\
0.28 & 69  & 100.00  & 100.00  &  100.00  & 100.00 \\
0.42 & 45  & 100.00  & 100.00  &  100.00  & 100.00 \\
\hdashline
0.57 & 62  & 100.00  & 100.00  &  100.00  & 100.00 \\
0.71 & 69  & 100.00  & 100.00  &  100.00  & 100.00 \\
0.85 & 76  & 100.00  & 100.00  &  100.00  & 100.00 \\
1.00 & 112 & 98.21   & 100.00  &  100.00  & 100.00 \\
\hline
\end{tabular}
\end{table}

As can be seen from these tables, for the BRAZILIAN dataset, the WI approach presented similar behavior to the other datasets (presented in the original article).
Thus, positive samples form a dense cluster (almost all positive samples have $IH \leq 0.14$), and the negative samples are scattered throughout space. The negative (random) samples may be disjoint to the positive set. The negative samples formed by the ``good quality skilled forgeries'' overlap the positive region of the DS, resulting in the need for a classifier with complex decision boundary.

\begin{table}[!htb]
\caption{Relationship between IH and accuracy (\%) for the negative (skilled) samples }
\label{tab:brazilian_accuracy_IH_negative_skilled_training_nn}
\scriptsize
\centering

\begin{tabular}{cccccc}
\hline
IH & \#Samples & R1 & R5$_{max}$ & R15$_{max}$ & R30$_{max}$\\ 
\hline

0.00 & 5   &  100.00 & 100.00  &  100.00  & 100.00 \\
0.14 & 9   &  100.00 & 100.00  &  100.00  & 100.00 \\
0.28 & 29  &  100.00 & 100.00  &  100.00  & 100.00 \\
0.42 & 23  &  100.00 & 100.00  &  100.00  & 100.00 \\
\hdashline
0.57 & 39  &  100.00 & 100.00  &  100.00  & 100.00 \\
0.71 & 63  &  100.00 & 100.00  &  100.00  & 100.00 \\
0.85 & 115 &  100.00 & 100.00  &  100.00  & 100.00 \\
\textbf{1.00} & \textbf{317} &  \textbf{93.05}  &  \textbf{95.89}  &  \textbf{98.42}   & \textbf{98.73} \\
\hline
\end{tabular}
\end{table} 

\newpage

\section*{S2 Complementary study on ``good'' and ``bad'' quality skilled forgeries}

\begin{figure}[!htb]
\centering
  \includegraphics[width=\columnwidth]{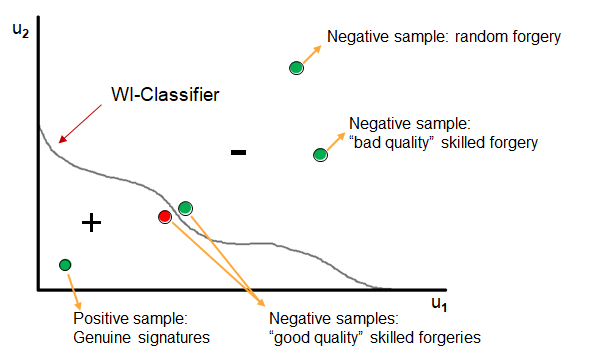}
  \caption{Synthetic decision frontiers. The same as in Figure \ref{fig:decision_frontier}  from the original article.}
  \label{fig:espaco}
\end{figure}

In this section we present a complementary study on the Figure \ref{fig:decision_frontier} of the original article, including the instance hardness analysis.
In addition, an image-level analysis is also carried out for an easier and better understanding of the scenarios. 
Figure \ref{fig:espaco} depicts the same behaviour as in Figure \ref{fig:decision_frontier} , highlighting key instances, which are:
\begin{itemize}
\item Positive sample: Genuine signature
\item Negative sample: Random forgery
\item Negative sample: ``Bad quality'' skilled forgery
\item Negative sample (correctly classified): ``Good quality'' skilled forgery
\item Negative sample (wrongly classified): ``Good quality'' skilled forgery
\end{itemize}

As presented in Figure \ref{fig:espaco}, while the negative region of space is located on the right of the decision boundary, the positive region is located on the left side. So, all correctly classified instances are colored in green, the wrongly classified one is presented in red.

\begin{figure}[!htb]
\advance\leftskip-2cm
  \includegraphics[]{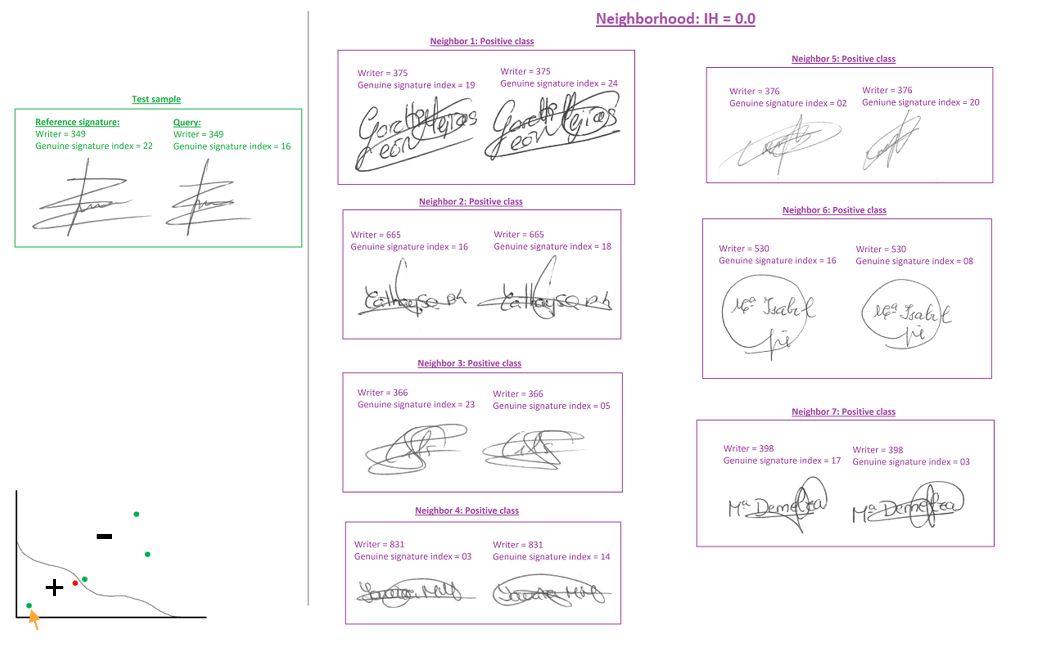}
  \caption{A positive tested sample on the left and its neighborhood on the right.}
  \label{fig:genuine_signature}
\end{figure}

\begin{figure}[!htb]
\advance\leftskip-2cm
  \includegraphics[]{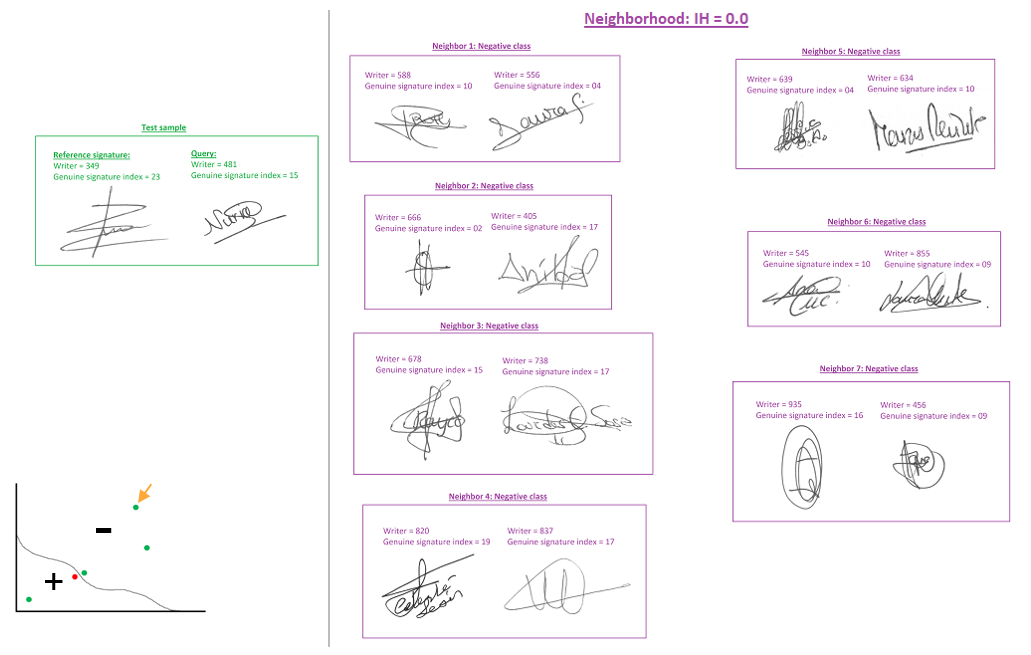}
  \caption{A negative tested sample (random forgery) on the left and its neighborhood on the right.}
  \label{fig:random_forgery}
\end{figure}

\begin{figure}[!htb]
\advance\leftskip-2cm
  \includegraphics[]{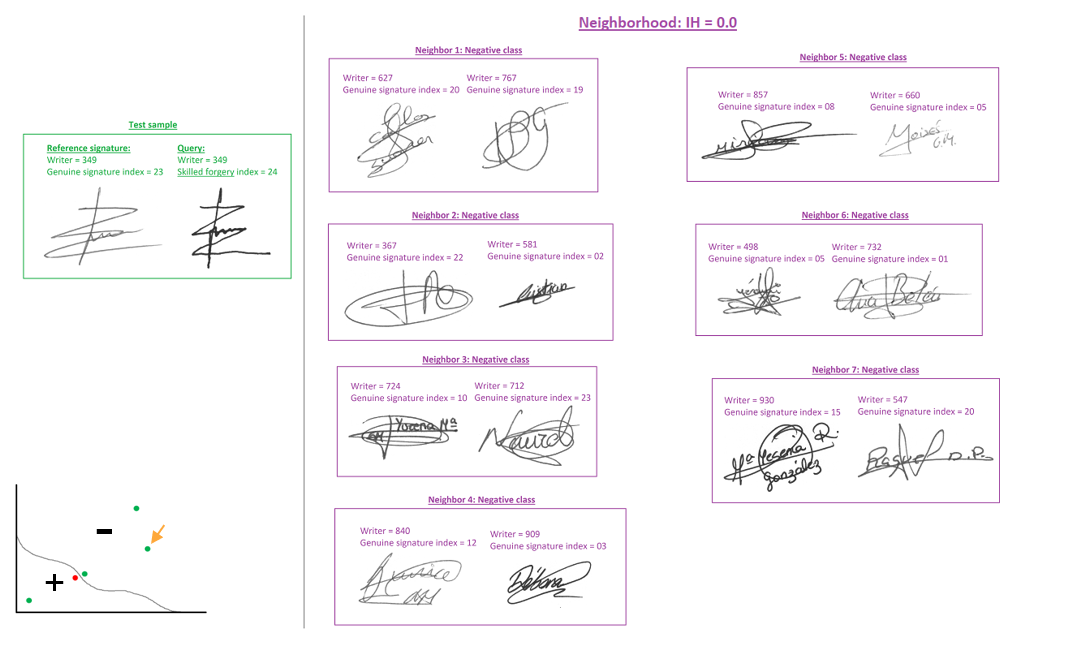}
  \caption{A negative tested sample (``Bad quality'' skilled forgery) on the left and its neighborhood on the right.}
  \label{fig:bad_quality}
\end{figure}

\begin{figure}[!htb]
\advance\leftskip-2cm
  \includegraphics[]{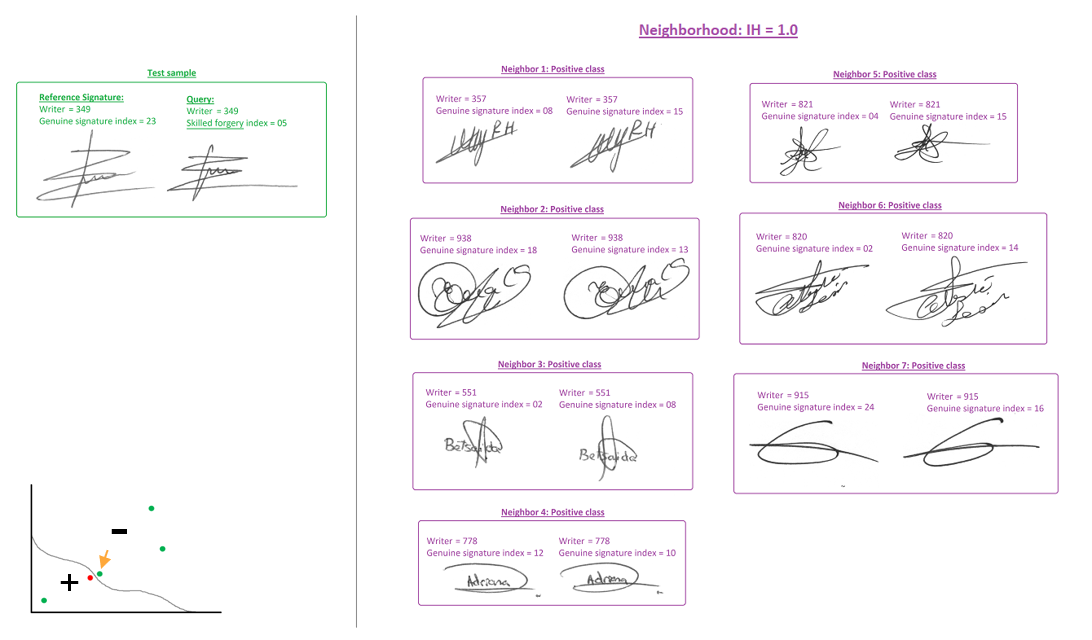}
  \caption{A negative tested sample correctly classified (``Good quality'' skilled forgery) on the left and its neighborhood on the right.}
  \label{fig:good_quality_correct}
\end{figure}

\begin{figure}[!htb]
\advance\leftskip-2cm
  \includegraphics[]{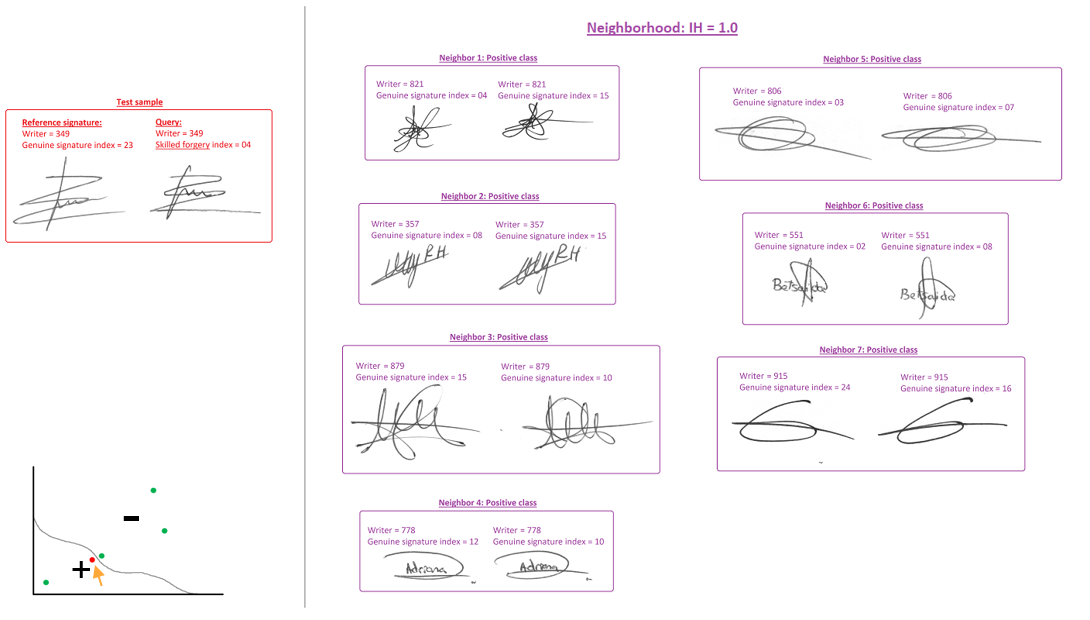}
  \caption{A negative tested sample wrongly classified (``Good quality'' skilled forgery) on the left and its neighborhood on the right.}
  \label{fig:good_quality_wrong}
\end{figure}

Figures \ref{fig:genuine_signature}, \ref{fig:random_forgery}, \ref{fig:bad_quality}, \ref{fig:good_quality_correct} and \ref{fig:good_quality_wrong} present, respectively, the behavior of genuine signature, random forgery, ``bad quality'' skilled forgery, ``good quality'' skilled forgery (correctly classified) and ``good quality'' skilled forgery (wrongly classified) at the image level for the GPDS dataset \cite{bonilla:07}. 

In all these figures, while left side presents the tested sample, the right side (in purple) contains the neighborhood of the tested sample in the training set (same methodology as in the original article). Recall, to obtain a dissimilarity vector we need a reference signature and a query. That is the reason is the reason why in each sample two signatures are presented. Each sample, also contains the index of the writer and the index of the signature.

On the top of each figure, the instance hardness value related to the tested sample is presented. Also, in the lower left corner of each figure, the location of the respective instance in Figure \ref{fig:espaco}.

\newpage

As depicted in Figure \ref{fig:genuine_signature}, the tested positive sample is formed by two genuine signatures from the same writer (index 349). As can be seen, all instances of the neighborhood belong to the positive class, since both signatures used to obtain the dissimilarity vector are from the same writer. So, the $IH = 0.0$. 
Finally, as both references and queries are formed by similar signatures all these dissimilarity vectors are located close to the origin (as highlighted in lower left corner of the figure).


\newpage

As depicted in Figure \ref{fig:random_forgery}, the tested negative sample is formed by two signatures from different writers (index 349 and 481), which clearly has different formats. 
The same behavior can also be seen in all instances that belong to the neighborhood. As all neighbors belong to the negative class, then $IH = 0.0$. 
Finally, as both references and queries are formed by signatures from different writers and have different formats all these dissimilarity vectors are located far from the origin (as highlighted in lower left corner of the figure).

\newpage

As depicted in Figure \ref{fig:bad_quality}, the tested negative sample is formed by genuine signature as reference and a ``bad quality'' skilled forgery. On the eye, one can clearly see that forgery is not good.
Thus, the ``bad quality'' skilled forgery behaves similarly to a random forgery in the dissimilarity space. So that, as can be seen, all neighbors belong to the negative class and are formed by dissimilarity vectors formed by signatures of different writers and different formats ($IH = 0.0$). As highlighted in lower left corner of the figure, it is located far from the origin.

\newpage

As presented in Figures \ref{fig:good_quality_correct} and \ref{fig:good_quality_wrong}, a ``good quality'' skilled forgery actually looks like the genuine signature. Thus, the ``good quality'' skilled forgery behaves similarly to a genuine signature in the dissimilarity space. So that, as can be seen, all neighbors belong to the positive class and are formed by dissimilarity vectors formed by signatures from the same writers ($IH = 1.0$).

The fact of being located close to the WI decision boundary results in hard to classify instances. Consequently, correct (Figure \ref{fig:good_quality_correct}) and wrong (\ref{fig:good_quality_wrong}) classification may occur for this type of test samples.

\end{document}